\documentclass{article}
\usepackage{times}
\usepackage{graphicx}
\usepackage{algorithm2e}
\usepackage{xcolor}
\usepackage{subcaption}
\usepackage{amsmath, amsfonts, amssymb}
\usepackage{tabularx}
\usepackage[table]{xcolor}
\usepackage{booktabs}
\usepackage{multirow}
\usepackage{amsthm}
\usepackage{enumitem}
\newtheoremstyle{normal} 
  {6pt}                  
  {6pt}                  
  {\normalfont}          
  {}                     
  {\bfseries}            
  {.}                    
  {.5em}                 
  {}                     
\theoremstyle{normal}
\newtheorem{theorem}{Theorem}
\usepackage{mathrsfs}
\usepackage{setspace}
\usepackage[font=small]{caption}
\usepackage{threeparttable}
\usepackage[normalem]{ulem}
\usepackage{wrapfig}

\newcommand{\name}{{\textit{ExtremControl}\xspace}}

\newcommand{\cyan}[1]{\textcolor[RGB]{64,255,165}{#1}}
\newcommand{\orange}[1]{\textcolor[RGB]{255,165,64}{#1}}
\newcommand{\purple}[1]{\textcolor[RGB]{165,64,255}{#1}}

\newcommand{\cmark}{\textcolor{green!60!black}{$\checkmark$}}
\newcommand{\xmark}{\textcolor{red!70!black}{$\times$}}

\usepackage[final]{corl_2026} 

\title{ExtremControl: Low-Latency Humanoid Teleoperation with Scaled Velocity Feedforward}

\author{
  Ziyan Xiong$^{1*\dagger}$ \quad Lixing Fang$^{1*}$ \quad Junyun Huang$^{1*}$ \quad Kashu Yamazaki$^{2}$ \\
  \bfseries Hao Zhang$^{1}$ \quad Chuang Gan$^{1}$ \\[2pt]
  {\normalfont $^{1}$University of Massachusetts Amherst \quad $^{2}$Carnegie Mellon University} \\
  {\normalfont $^{*}$Equal contribution \quad $^{\dagger}$Corresponding author: \texttt{ziyanxiong@umass.edu}}
}

\begin{document}
\maketitle

\begin{center}
    \centering
    \captionsetup{type=figure}
    \vspace{-19pt}
    \includegraphics[width=1.0\textwidth]{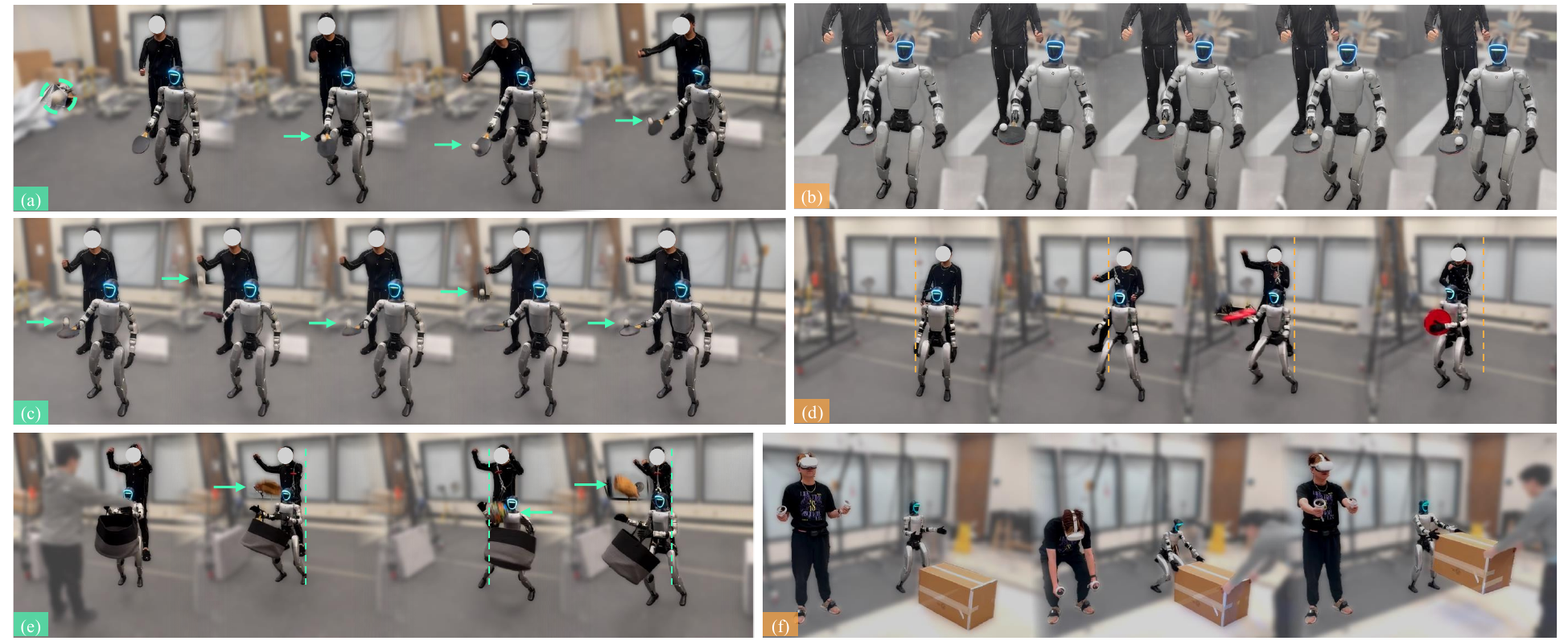}
    \caption{Unitree G1 demonstrates a diverse set of loco-manipulation tasks under teleoperation: (a) returning a ping-pong ball from varying positions; (b) balancing a ping-pong ball on a paddle through rapid orientation adjustments; (c) juggling a ping-pong ball; (d) catching a frisbee while moving; (e) catching thrown objects using a handheld basket while in motion; and (f) cooperatively lifting a box. Tasks (a-e) are performed using an optical MoCap system to achieve lower latency, while task (f) is operated using a VR system.}
\label{fig:teaser}
\end{center}

\begin{abstract}
Building a low-latency humanoid teleoperation system is essential for collecting diverse reactive and dynamic demonstrations. However, end-to-end latencies in existing approaches consistently hover around 200\,ms, roughly equal to human reaction time, that severely limits responsiveness and prevents tasks requiring rapid feedback. We identify position-only PD control as the root cause of this latency floor and propose \name, a low-latency whole-body control framework that: (1) incorporates scaled velocity feedforward in low-level control, reducing tracking delay by approximately 70\,ms with a provably optimal feedforward ratio; (2) operates directly on $\mathrm{SE}(3)$ poses of selected rigid links, primarily humanoid extremities, to avoid joint-space retargeting in the control loop. To minimize retargeting time, we derive a closed-form Cartesian-space mapping that uses a single anthropomorphic calibration step to translate human motion into humanoid link targets, accommodating operators with varying body proportions. Furthermore, we provide a theoretical framework for selecting control hyperparameters, and validate the effectiveness of velocity feedforward control in the reinforcement learning for locomotion setting through experiments in both simulation and the real world. Building on \name, we implement a teleoperation system supporting both optical motion capture and VR-based tracking, achieving end-to-end latency as low as 50\,ms --- well beneath the threshold at which system delay adds a full human reaction cycle to the control loop. This enables a qualitatively new class of reactive behaviors, including ping-pong ball balancing, juggling, and real-time return, that are beyond the capability of existing teleoperation systems. Project page: \url{https://extremcontrol.github.io/}.
\end{abstract}


\clearpage

\section{Introduction} 
\begin{table*}[t]
\centering
\begin{threeparttable}
\setlength{\tabcolsep}{6pt}
\renewcommand{\arraystretch}{1.4}
\setlength{\aboverulesep}{2pt}
\setlength{\belowrulesep}{-1pt}
{\scriptsize
\begin{tabular*}{\textwidth}{l|c|c|c|c|c}
\toprule
\\[-10pt]
\rowcolor{gray!15}
\textbf{Teleoperation System} & 
\textbf{Control Interface} & \textbf{Wrist Control} & \textbf{Foot Control} & \textbf{Task-Space Retargeting}\hspace{-4pt} & \textbf{End-to-End Latency}
\\
\hline
HOMIE \cite{HOMIE} & Exoskeleton & \cmark & \xmark & \cmark & $\sim454$\,ms
\\[-2pt]
HumanPlus \cite{HumanPlus} & RGB Camera & \cmark & \cmark & \xmark & $\sim340$\,ms
\\[-2pt]
OmniH2O \cite{omnih2o} & VR & \cmark & \xmark & \cmark &
$\sim$ \textbf{185\,ms}
\\[-2pt]
H2O \cite{h20} & RGB Camera & \xmark & \cmark & \xmark & $\sim373$\,ms
\\[-2pt]
AMO \cite{AMO} & VR & \cmark & \xmark & \xmark & $\sim 380$\,ms
\\[-2pt]
CLONE \cite{CLONE} & VR & \cmark & \xmark & \cmark & $\sim$ \textbf{178\,ms}
\\[-2pt]
AMS \cite{AMS} & MoCap & \xmark & \cmark & \xmark & $\sim201$\,ms
\\[-2pt]
TWIST \cite{TWIST} & MoCap & \xmark & \cmark & \xmark & $>700$\,ms
\\[-2pt]
TWIST2 \cite{TWIST2} & VR & \cmark & \cmark & \xmark & $\sim234$\,ms
\\[-2pt]
SONIC \cite{SONIC} WBC & RGB Camera, VR & \cmark & \cmark & \xmark & $\sim427$\,ms\tnote{a}
\\[-2pt]
SONIC \cite{SONIC} Keypoint & VR & \cmark & \xmark & \cmark & ---\tnote{b}
\\[-2pt]
ExtremControl (Ours) & VR, MoCap & \cmark & \cmark & \cmark & $\sim$ \textbf{54\,ms}
\\[-2pt]
\bottomrule
\end{tabular*}
}
\begin{tablenotes}
\scriptsize
\item[a] Measured under VR setting. RGB camera will introduce 0.1\,s additional latency.
\item[b] Latency not measurable as the teleoperator is not visible in the released video.
\end{tablenotes}
\end{threeparttable}
\vspace{0pt}
\caption{Existing humanoid teleoperation systems with latencies estimated by optical flow on released videos.}
\label{table:existing-systems}
\vspace{-15pt}
\end{table*}

Humanoid robots have long attracted significant attention in the robotics community due to their human-like morphology and kinematic structure. Because modern environments, tools, and tasks are predominantly designed around human bodies, humanoids represent a natural embodiment for general-purpose robotic systems capable of operating in unstructured, human-centric settings. Moreover, the close correspondence between humanoid and human morphology enables the direct exploitation of large-scale human motion and skill datasets, alleviating the reliance on expensive and limited robot-collected data. However, these same properties that make humanoids appealing also pose substantial challenges for traditional control frameworks. In particular, the high-dimensional state and action spaces, underactuated floating-base dynamics, intermittent contacts, and frequent hybrid mode transitions inherent to humanoid locomotion and manipulation render classical model-based control approaches difficult to scale and deploy robustly in practice. Accurate modeling, real-time optimization, and contact-consistent planning become increasingly intractable as task complexity grows, motivating the exploration of alternative control paradigms better suited to the complexity of humanoid systems.

With the advent of large-scale parallel simulation, reinforcement learning has become a dominant paradigm for humanoid locomotion and whole-body control, and the design of control interfaces has undergone a clear evolution over time. Inheriting from quadruped locomotion, early approaches relied on explicit Cartesian-space objectives \cite{omnih2o, h20} or command-based interfaces \cite{WoCoCo, humanoidparkourlearning}. Subsequently, influenced by advances in physics-based character animation \cite{PHC, AMP}, many methods shifted toward dense joint-space pose supervision adopted from the human model, enabling humanoids to reproduce expressive and highly dynamic motions derived from human demonstrations \cite{HumanPlus, HOVER}. More recent efforts further narrowed the control objective to optimization of selected reference trajectories, achieving high-fidelity motion reproduction at the cost of generalization across tasks and behaviors \cite{videomimic,GMT,ASAP,beyondmimic}. In parallel, teleoperation-oriented methods introduce intermediate representations to flexibly map human motion to humanoid embodiments \cite{AMO,AMS,TWIST,TWIST2}. These approaches first compute target poses for a set of robot links and then solve inverse kinematics to obtain joint configurations that realize those poses. However, although policies are executed in joint space, the underlying optimization objective remains defined in terms of matching Cartesian link poses.

A large portion of the literature on humanoid control focuses on teleoperation \cite{HOMIE, HumanPlus, omnih2o, h20, AMO, CLONE, AMS, TWIST, TWIST2}, as it provides an effective mechanism for collecting data to train general-purpose robotic intelligence. Because teleoperation operates in a human-in-the-loop closed-loop setting, system latency critically determines the operator’s ability to perform responsive tasks. Among the existing teleoperation systems listed in Tab.~\ref{table:existing-systems}, we observe a striking consistency: \textbf{all existing humanoid teleoperation systems exhibit end-to-end latencies over 150\,ms}, independent of the robot, retargeting strategy, or the length of future motion used \cite{omnih2o, CLONE, AMS, TWIST2}.\footnote{We estimate latencies by our proposed optical flow method on authors' released videos.} 
We identify position-only PD control as a common contributing factor to this latency floor: without a velocity feedforward term, the low-level controller introduces an inherent tracking delay that the learned policy cannot fully compensate for, particularly in the teleoperation setting where future reference is limited and rapid response is most critical.


To address this problem, we introduce \name, a humanoid whole-body control framework that: (1) incorporates a scaled velocity feedforward term for extremely responsive low-level actuation, supported by a whole-body impedance calibration that tightly couples simulation with real-world deployment and provides a principled basis for hyperparameter selection; and (2) utilizes a closed-form Cartesian-space mapping with one-shot anthropomorphic calibration to convert human motion to $\mathrm{SE}(3)$ poses of selected rigid links spanning all humanoid extremities, generalizing across operators of varying body proportions without requiring a human shape prior. The overall design aims to minimize system latency while preserving full whole-body control capability.

The contributions of this work are fourfold. First, we present a theoretical framework for velocity feedforward control in the reinforcement learning for locomotion setting, deriving the relationship between the feedforward ratio $\eta$ and tracking delay, establishing a principled upper bound on $\eta$ to prevent discrete-time overshoot, and providing whole-body impedance calibration as a systematic procedure for PD gain selection; we validate this framework through experiments in both simulation and the real world. Second, we introduce Direct Extremity Control, a retargeting-free Cartesian-space mapping that achieves higher end-effector fidelity than full-body joint-space retargeting while requiring only a one-shot calibration and supporting operators of varying body proportions. Third, we propose an optical-flow-based latency estimation method that measures end-to-end teleoperation delay directly from video without requiring internal timestamps, enabling fair cross-system comparison from publicly released demonstrations. Finally, leveraging \name, \textbf{we develop a humanoid teleoperation system achieving as low as 50\,ms end-to-end latency}, well beneath the threshold at which system delay adds a full human reaction cycle to the control loop, thereby unlocking a qualitatively new class of reactive loco-manipulation tasks, including ping-pong ball balancing, juggling, and real-time return (Fig.~\ref{fig:teaser}), that lie beyond the reach of existing teleoperation systems.


\section{Related Work}


\paragraph{Reinforcement Learning for Locomotion} Reinforcement learning has become a dominant paradigm for learning agile locomotion policies, due to its ability to optimize high-dimensional control objectives directly from interaction. Early work demonstrated that model-free RL can produce robust locomotion behaviors in simulation and transfer them to real robots through domain randomization and privileged training signals \cite{LearningRobustRewards, RMA, leggedgym, Sim-to-Real, xiong2024mqe, fang2026spotspatialperceptionorientedlonghorizon}. Subsequent studies improved robustness and versatility by introducing curriculum learning, command-conditioned policies \cite{HiFAR, ExpressiveWBCHumanoid, Hwangbo2019learningagileanddynamic, walktheseways, LearningToWalk, robotparkourlearning}. Hybrid approaches further combine trajectory optimization, model-based priors, or analytical controllers with RL fine-tuning to enhance stability and tracking accuracy \cite{ResidualRLRobotControl, HoldMyBeer, ContinuousVersatileJumping, ImitatingAndFinetuningQuadrupedal}. Together, these works establish RL as a practical framework for locomotion across diverse tasks and robot embodiments.

\paragraph{Humanoid Whole-Body Control} Compared to quadrupeds, humanoid locomotion and loco-manipulation involve high degrees of freedom, underactuated contacts, and complex whole-body coordination. Whole-body control (WBC) frameworks address these challenges by tracking joint-space or task-space objectives under full-body dynamics and contact constraints~\cite{ExpressiveWBCHumanoid, AtlasFullBodyControl, PassiveBasedTorqueControl, TorqueControllHumanoid, humanoidWBCFramework, SYNTHESISWBC}.
Recent approaches integrate reinforcement learning with WBC, using motion capture and animation data to provide expressive reference motions~\cite{AMS,SONIC,PHC,AMP,KungfuBot2,bfmzero,lv2025lafan1_retargeting,HumanoidPose,omniretarget,HuB}. 
Humanoid teleoperation, which is a fundamental mechanism for large-scale data collection, fits naturally into this paradigm. Early methods specify Cartesian objectives for selected body links~\cite{HOMIE,omnih2o,CLONE,CHILD,HeavyLifting,ACE} with control interfaces ranging from exoskeleton to single rgb camera; whereas recent approaches relied on full-body joint-space retargeting, incurring added latency~\cite{h20,TWIST,videomimic,GMR,MobileTeleVision,SPARK}.
Overall, WBC serves as a unifying backbone for complex humanoid systems.


\newpage
\section{Method}
\subsection{Velocity Feedforward Control}

\label{sec:ffcontrol}

\paragraph{Whole-Body Impedance Calibration} To facilitate a principled design of joint-level PD controllers, we calibrate gains using a sequential, simulation-based procedure that estimates effective joint impedance under whole-body closed-loop control. For each joint, the derivative gain is temporarily set to zero and the proportional gain is sampled as $k_p^{(e)} \sim \mathcal{U}(0.5k_p, 1.5k_p)$ across parallel simulation environments $e \in E$. Applying a small perturbation and measuring the induced oscillation period $P^{(e)}$, the effective joint impedance is estimated as (see Sec.~\ref{sec:proof_impedance} for proof)
\begin{equation}
M_{\mathrm{eff}}^{(e)} = \frac{k_p^{(e)} \left(P^{(e)}\right)^2}{(2\pi)^2}.
\end{equation}
The average effective joint impedance $\overline{M_{\mathrm{eff}}}$ over different environments is then used to set the PD gains given a target natural frequency $\omega_n$ and damping ratio $\zeta$:
\begin{equation}
k_p := \overline{M_{\mathrm{eff}}}\,\omega_n^2, \qquad k_d := 2\zeta\,\overline{M_{\mathrm{eff}}}\,\omega_n,
\label{eq:pd_from_impedance}
\end{equation}
This procedure iterates over all joints from distal to proximal, yielding dynamically consistent gains that serve as a stable and well-scaled initialization for downstream control and learning.

\paragraph{Scaled Velocity Feedforward}
With velocity feedforward ratio $\eta \in [0,1]$, the PD control becomes
\begin{equation}
\tau = k_p(q^* - q) + k_d(\eta\,\dot{q}^* - \dot{q}).
\end{equation}

Prior work has shown that position targets outperform velocity targets when used independently as policy actions~\cite{2017-SCA-action}. Due to safety considerations, position-based action spaces have become the predominant choice in reinforcement learning for locomotion. Our method retains position targets as policy outputs while approximating velocity targets through finite differences of consecutive policy actions as
\begin{equation}
\dot{q}^*(t) = \frac{q^*(t) - q^*(t - \Delta t)}{\Delta t}.
\end{equation}
This allows $\eta$ to be tuned as a continuous knob between pure position control ($\eta=0$) and full velocity feedforward ($\eta=1$). The following two theorems characterize the effects of $\eta$ on tracking delay and control stability, providing a principled basis for selecting the velocity feedforward ratio (proofs in Sec.~\ref{sec:proof}).
\begin{theorem}
\label{thm:one}
Under whole-body impedance calibration with critical damping $\zeta=1$, the tracking delay $\ell_\mathrm{control}$ for a sinusoidal target $q^*=A\sin(\omega t)$ satisfies
\begin{equation}
\ell \approx \frac{2(1-\eta)}{\omega_n}, \quad \omega \ll \omega_n.
\label{eq:latency}
\end{equation}
\end{theorem}
When $\eta < 1$, one might expect the policy to implicitly compensate for this delay by learning to output joint positions that are offset from the true targets, effectively anticipating future states. However, we will show in Sec.~\ref{sec:experiment} that such learned compensation cannot fully eliminate the delay in the teleoperation setting where responsiveness is required.
\begin{theorem}
\label{thm:two}
Under zero-order hold with control interval $\Delta t$, to prevent overshoot caused by excess acceleration within a single control interval, the feedforward ratio must satisfy
\begin{equation}
\eta \le 1 - \frac{\omega_n \Delta t}{4}.
\label{eq:upperbound}
\end{equation}
\end{theorem}
The standard formulation $\eta=1$ is therefore suboptimal, and the admissible range of $\eta$ shrinks with higher natural frequency $\omega_n$ or longer control interval $\Delta t$.

\subsection{Direct Extremity Control}
\label{sec:method_retarget}

\begin{wrapfigure}{r}{0.5\textwidth}
    \centering
    \includegraphics[width=0.5\textwidth]{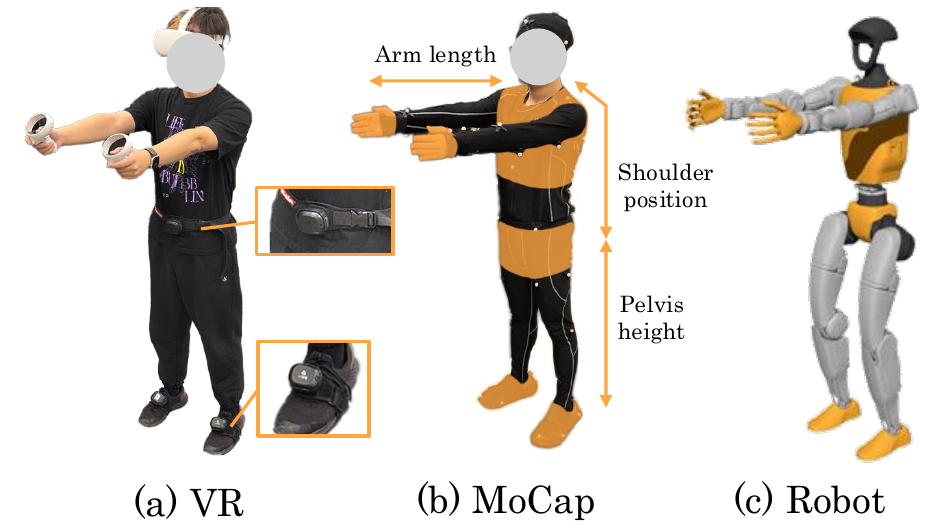}
    \caption{\textbf{\orange{Tracking objectives}} for humans and humanoids under VR and MoCap settings.}
    \label{fig:calibration}
    \vspace{-15pt}
\end{wrapfigure}

\paragraph{Tracking Objectives}
The choice of tracking objectives fundamentally determines the feasibility and robustness of a teleoperation system. Under the contact-sparse assumption that covers most practical scenarios, we select six links of the Unitree G1, which are hands, feet, torso, and pelvis, yielding $\mathbf{T}^r \in \mathrm{SE}(3)^6$. These six links are sufficient to convey the operator's intent for manipulation, locomotion, and global posture, while remaining minimal enough to avoid over-constraining the system. Although the Cartesian pose of a hand does not uniquely determine the arm's joint configuration, joint limits, self-collision constraints, and temporal continuity strongly restrict the feasible solution space. We further demonstrate in Sec.~\ref{sec:experiment} that extremity poses alone achieve performance comparable to adding retargeted joint configurations, while greatly surpassing retargeted joint configurations only.

\paragraph{Mapping Design}
We define a Cartesian-space mapping $\mathcal{M}: \mathrm{SE}(3)^6 \rightarrow \mathrm{SE}(3)^6$ following three principles: (i) \emph{anthropomorphic compensation}: explicitly handling differences in body proportions; (ii) \emph{mathematical consistency}: reproducing identical poses at calibration; and (iii) \emph{real-time efficiency}: avoiding multi-link or joint-space optimization.

\paragraph{One-Shot Calibration}
At initialization, the human stands in a neutral pose as shown in Fig.~\ref{fig:calibration}. Per-link rotational offsets are estimated to resolve coordinate-system differences and held fixed at runtime. A small set of anthropomorphic measurements is extracted from $\mathbf{T}^h$: pelvis height $z^h_{\text{pelvis}}$, shoulder positions in the pelvis frame $\mathbf{p}^h_{k\text{\_shoulder}}$, and arm lengths $l^h_{k\text{\_arm}}$ for each side $k \in \{\text{left}, \text{right}\}$. Corresponding humanoid parameters $\{z^r_{\text{pelvis}}, \mathbf{p}^r_{k\text{\_shoulder}}, l^r_{k\text{\_arm}}\}$ are read directly from the kinematic model. A fixed foot offset $\delta_{k\text{\_foot}} = \mathbf{p}^r_{k\text{\_foot}} - ({z^r_{\text{pelvis}}}/{z^h_{\text{pelvis}}}) \cdot \mathbf{p}^{h}_{k\text{\_foot}}$ is also recorded.
At runtime, humanoid targets $\mathbf{T}^r = \mathcal{M}(\mathbf{T}^h)$ are computed as follows. 

\textbf{Pelvis and feet} are height-ratio scaled:
\begin{equation}
    \mathbf{p}^r_{\text{pelvis},k\text{\_foot}} = \frac{z^r_{\text{pelvis}}}{z^h_{\text{pelvis}}} \cdot \mathbf{p}^h_{\text{pelvis},k\text{\_foot}}.
\end{equation}
\textbf{Torso} position is rigidly attached to the pelvis via the fixed kinematic offset $\mathbf{p}_{\text{diff}}$:
\begin{equation}
    \mathbf{p}^r_{\text{torso}} = \mathbf{p}^r_{\text{pelvis}} + \mathbf{R}^r_{\text{pelvis}}\,\mathbf{p}_{\text{diff}}.
\end{equation}
\textbf{Hands} are mapped by aligning shoulder positions and scaling arm lengths in the torso frame:
\begin{equation}
    \frac{\mathbf{T}^r_{k\text{\_hand}}(\mathbf{T}^r_{\text{torso}})^{-1} - \mathbf{p}^r_{k\text{\_shoulder}}}{l^r_{k\text{\_arm}}} = \frac{\mathbf{T}^h_{k\text{\_hand}}(\mathbf{T}^h_{\text{torso}})^{-1} - \mathbf{p}^h_{k\text{\_shoulder}}}{l^h_{k\text{\_arm}}}.
\end{equation}
The entire mapping consists solely of rigid transformations and is computed in closed form making it fully feedforward and parallelizable.

\paragraph{Policy Learning} We adopt an RL+BC+RL three-stage framework to learn a deployable whole-body tracking policy. In the first stage, we train a teacher policy $\pi_{\mathrm{teacher}}$ via PPO~\cite{PPO} with privileged observations, including retargeted joint configurations from GMR~\cite{GMR}, global pose discrepancies, domain randomization parameters, and foot contact forces over a long future horizon of 32 frames. In the second stage, we distill $\pi_{\mathrm{teacher}}$ into a student policy $\pi_{\mathrm{student}}$ via DAgger~\cite{DAGGer}, removing all privileged observations (including the retargeted joint configurations) and reducing the future horizon to a single frame; this inevitably introduces a performance gap as the student loses access to the joint-space retargeting signal. The third stage recovers this capability by fine-tuning $\pi_{\mathrm{student}}$ with PPO over a large, diverse motion dataset to obtain $\pi_{\mathrm{teleop}}$, where online exploration is stabilized by freezing the actor for the first 200 iterations and annealing the entropy coefficient toward zero. Please refer to Sec.~\ref{sec:policy_learning} for policy learning details.

\section{Evaluation}

\label{sec:experiment}

\begin{figure}[t]
    \centering
    \includegraphics[width=0.8\textwidth]{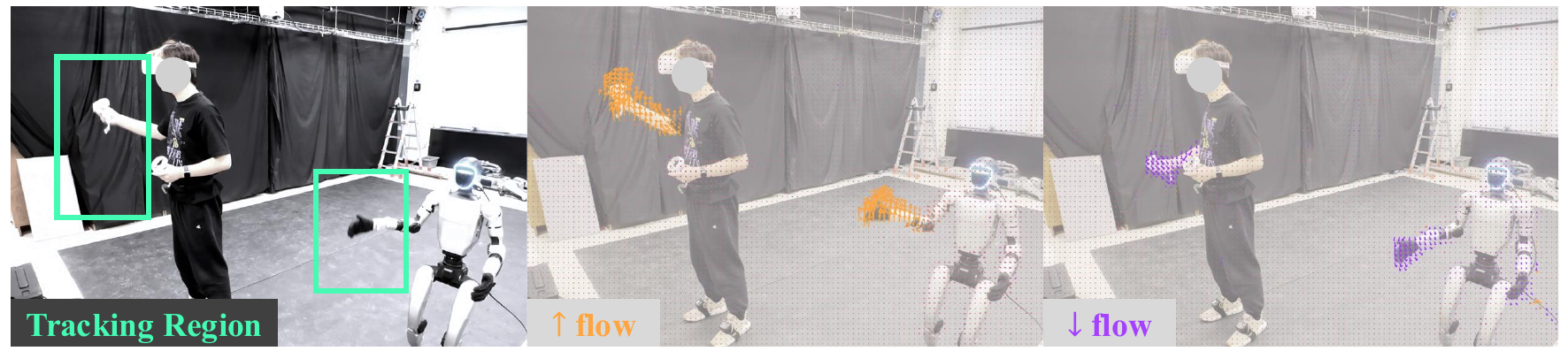}
    \caption{\textbf{Optical flow visualization of hand movements.} \orange{Upward} and \purple{downward} optical flow extracted from video clips are visualized in the two rightmost images. Separate tracking regions are defined for the human and robot hands to isolate their respective motion. The average optical flow magnitude within each region serves as a proxy for hand velocity, as it is approximately proportional to the hand's image-plane speed.}
    \label{fig:optical_flow_setup}
    \vspace{-10pt}
\end{figure}


\begin{wrapfigure}{r}{0.38\textwidth}
    \vspace{-12pt}
    \centering
    \includegraphics[width=0.38\textwidth]{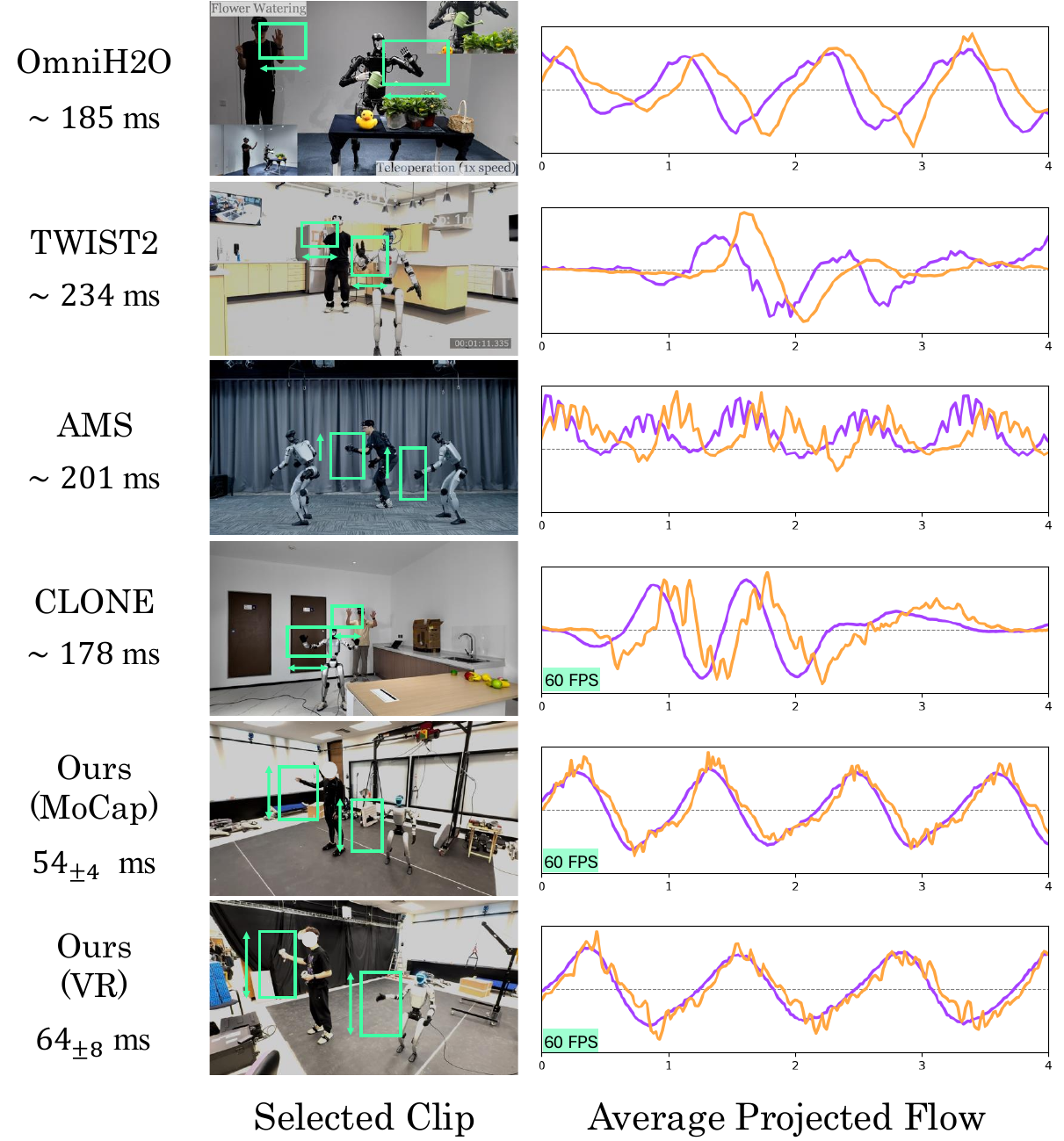}
    \vspace{-15pt}
    \caption{\textbf{\purple{Human} and \orange{robot} optical flow trajectories}, along with the corresponding system latency estimates derived from selected video clips of exsisting systems.}
    \label{fig:latency_analysis}
    \vspace{-10pt}
\end{wrapfigure}

To measure end-to-end latency without requiring internal timestamps, we develop a video-based optical flow method that captures externally observable cumulative latency across all system components. As illustrated in Fig.~\ref{fig:optical_flow_setup}, we define tracking regions on both the human operator and the robot, compute dense optical flow~\cite{farneback2003two} between consecutive frames, and project averaged flow vectors onto a predefined motion axis to yield a one-dimensional motion signal per region. A deliberate reciprocating hand motion produces a quasi-periodic signal for robust temporal alignment, and latency is estimated as the temporal offset that maximizes waveform alignment between the two signals. The same procedure is applied to public video demonstrations of third-party systems to enable fair comparison. Under this metric, \name~achieves 54\,ms end-to-end latency under the wired MoCap setup and 64\,ms under the wireless VR setup, outperforming all baselines as shown in Tab.~\ref{table:existing-systems} and Fig.~\ref{fig:latency_analysis}.

The correctness of \textbf{whole-body impedance calibration}, theoretical results of \textbf{velocity feedforward control}, and \textbf{end-to-end latency estimation} are validated through experiments in both simulation and real world. These experiment results as well as all experiment setups are detailed in Sec.~\ref{sec:additional_evaluation}. Further experiments address the following questions:

\orange{\textbf{Q1:} \textbf{How does Velocity Feedforward Control affect the policy performance?}}

From a control perspective, any desired joint torque $\tau^*(t)$ can theoretically be achieved under different velocity feedforward ratios $\eta$ by policy predicted target joint positions $q^*_\eta(t)$ that satisfy
\begin{align}
    k_p q^*_\eta(t) + \eta k_d \frac{q^*_\eta(t) - q^*_\eta(t-\Delta t)}{\Delta t} = \tau^*(t).
\end{align}
However, this equivalence does not necessarily hold from a reinforcement learning perspective. To encourage smooth actions for sim-to-real
transfer, locomotion policies are commonly trained with a penalty on consecutive action differences, $|a_t - a_{t-1}|$. Although different
values of $\eta$ can achieve the same torque, they generally require different action sequences and therefore incur different smoothness
penalties. Consequently, $\eta$ changes the effective optimization objective and can lead to different learned behavior,
even when all reward terms and their weights are kept identical.

To systematically investigate the effect of $\eta$ on control performance,  we consider two settings: 
(1) \textbf{teleoperation}, where the policy receives only the current reference frame and must respond 
immediately without access to future motion; and (2) \textbf{motion tracking}, where the policy is conditioned on 32 future reference frames
spanning 0.64\,s, allowing it to anticipate upcoming motion. Within each setting, we independently train policies with different values of
 $\eta$, while keeping all other configurations identical, including the reward functions and their weights. During evaluation, 
 each policy uses the same $\eta$ as in its training. We train each policy with five different random seeds.

\begin{wrapfigure}{r}{0.55\textwidth}
    \vspace{-15pt}
    \centering
    \includegraphics[width=0.55\textwidth]{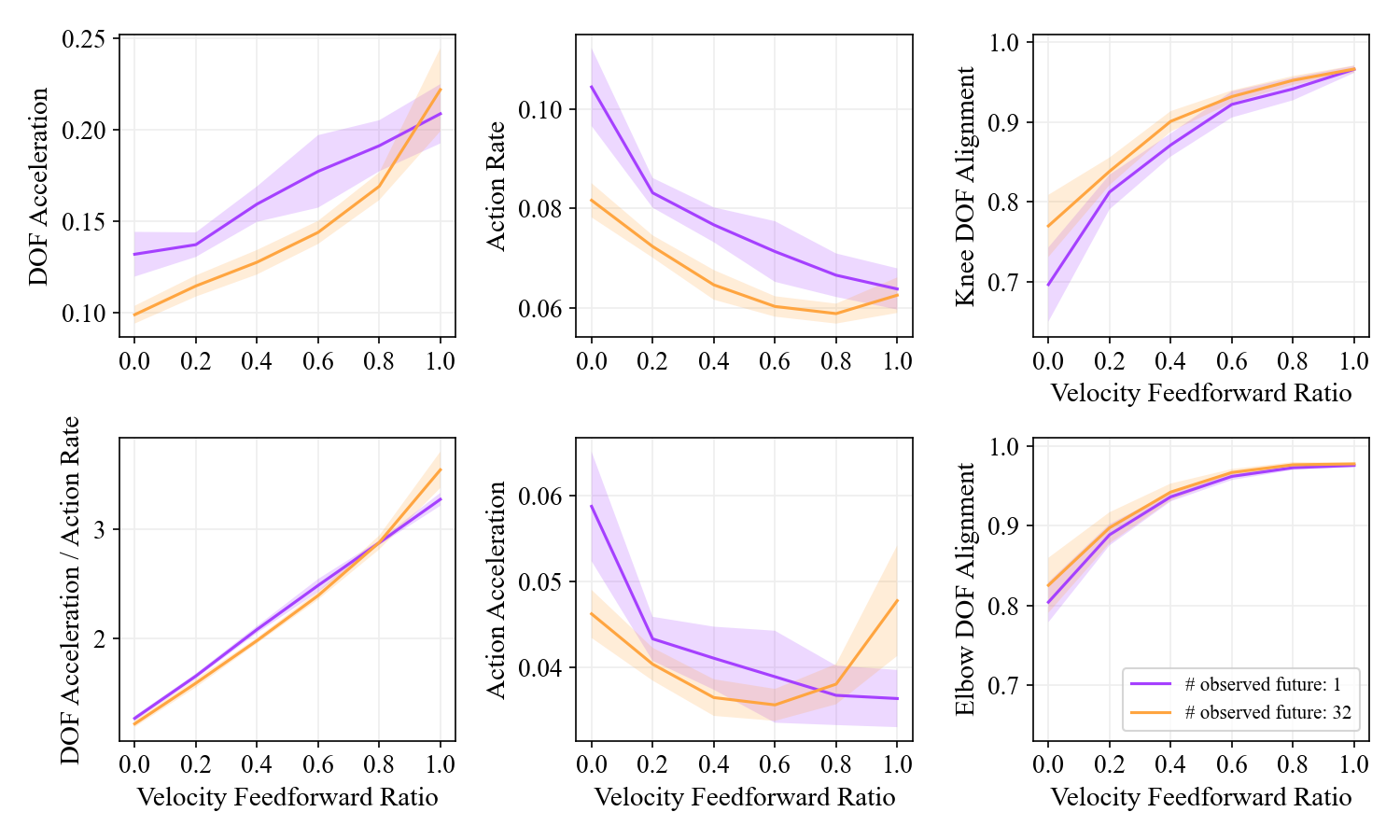}
    \vspace{-17pt}
    \caption{\textbf{Analysis of control characteristics under different velocity feedforward ratios.} We train \orange{motion tracking} and \purple{teleoperation} policies with varying velocity feedforward ratios and evaluate their control behavior across six metrics.}
    \label{fig:control_ablate}
    \vspace{-15pt}
\end{wrapfigure}
Fig.~\ref{fig:control_ablate} illustrates the control characteristics of policies trained with different velocity feedforward ratios.
Notably, in both teleoperation and motion tracking settings, the DoF acceleration per unit action rate (equivalently, torque per unit action rate)
increases approximately linearly with $\eta$, indicating greater control responsiveness to changes in policy actions.
Under a fixed action rate penalty $|a_t-a_{t-1}|^2$, this increased control responsiveness leads to higher measured joint accelerations, as shown
in upper-left panel of Fig.~\ref{fig:control_ablate}.

\begin{wrapfigure}{r}{0.40\textwidth}
    \vspace{-17pt}
    \centering
    \includegraphics[width=0.40\textwidth]{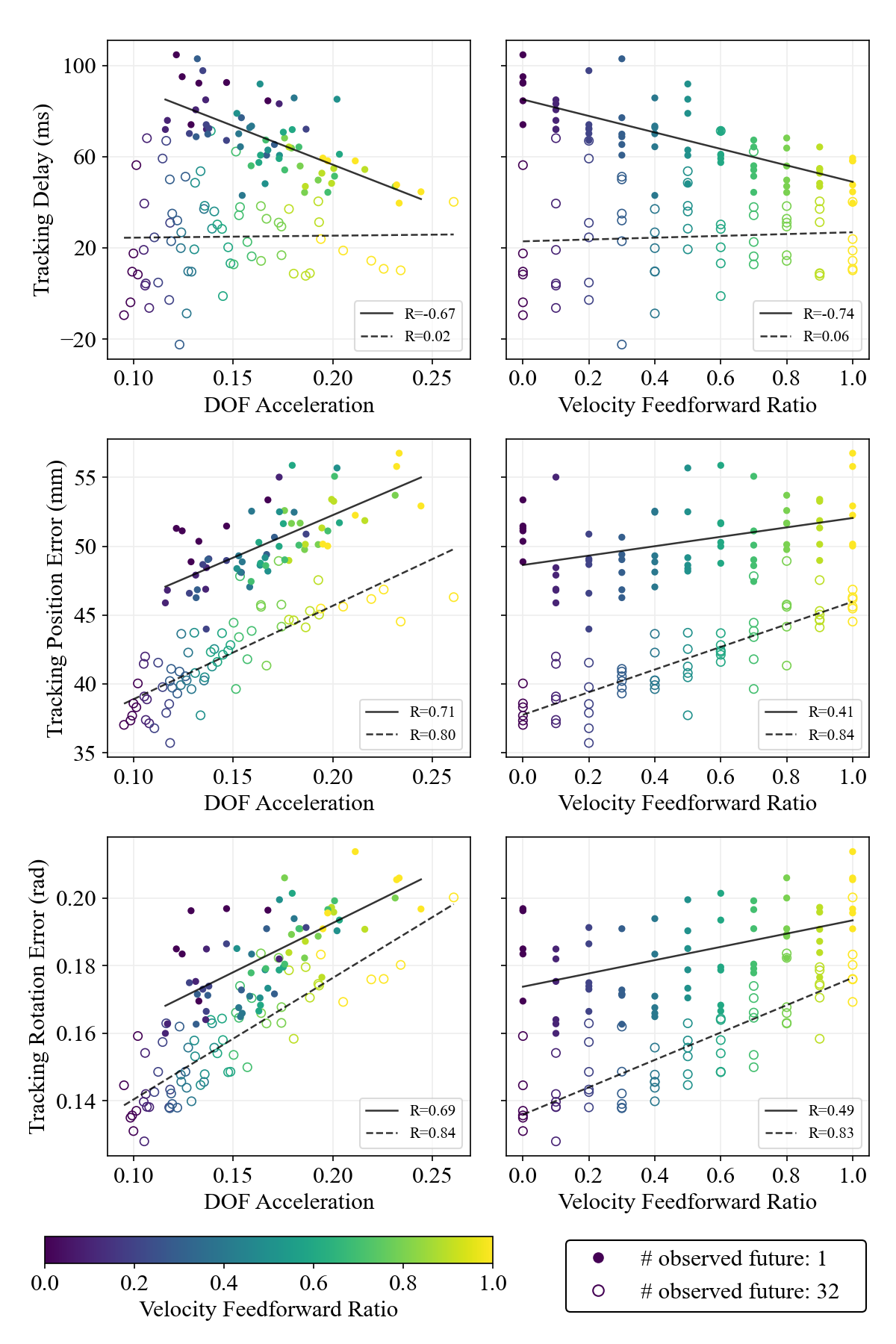}
    \vspace{-15pt}
    \caption{\textbf{Policy tracking performance with respect to different velocity feedforward ratio.} Each dot represents a policy.}
    \label{fig:latency_scatter}
    \vspace{-18pt}
\end{wrapfigure}

In \textbf{teleoperation} setting, as shown in the middle column of Fig.~\ref{fig:control_ablate}, \orange{both average action rate and action
acceleration consistently decrease as $\eta$ increases, indicating progressively smoother policy outputs}. This improved control responsiveness
also helps reduce tracking latency under the same action rate penalty, as demonstrated in Fig.~\ref{fig:latency_scatter}.
In contrast, in \textbf{motion tracking} setting, increasing $\eta$ initially reduces both the action rate and action acceleration. However,
excessively large $\eta$ leads to overly aggressive joint responses and potential overshooting, requiring the policy to make frequent corrective
adjustments. As a result, the action acceleration begins to increase first, followed by the action rate, revealing that access to future motion
references reduces the need for a large velocity feedforward ratio.

To investigate action interpretability, we measure the cosine similarity between policy actions and measured joint positions
(right column of Fig.~\ref{fig:control_ablate}). We consider both elbow and knee joints to capture non-contact and contact-rich settings,
respectively. Across both teleoperation and motion tracking settings, cosine similarity consistently increases with the velocity feedforward ratio $\eta$,
indicating that policy actions become more closely aligned with the actual joint positions.

Fig.~\ref{fig:latency_scatter} illustrates the tracking performance of policies with different velocity feedforward ratios, measured by tracking delay and tracking error,
in relation to the average measured DoF acceleration. Tracking delay is estimated as the temporal offset that best aligns the reference and measured end-effector trajectories (hand trajectories in our setting).
In \textbf{motion tracking} setting, the tracking delay remains approximately constant at 20\,ms (top row),
corresponding to a single policy action interval, regardless of the velocity feedforward ratio $\eta$. This suggests that access to future
motion references allows the policy to effectively compensate for low-level tracking delays, leaving little additional benefit from velocity
feedforward in terms of latency reduction. In contrast, in \textbf{teleoperation} setting, where future motion is unavailable,
\orange{tracking delay decreases approximately linearly as $\eta$ increases}, demonstrating the effectiveness of velocity feedforward
in improving control responsiveness without future motion information.

Tracking error consistently increases with the velocity feedforward ratio $\eta$ in both settings. However, tracking error exhibits a 
stronger linear correlation with the measured average DoF acceleration, as indicated by a higher regression $R^2$. As discussed earlier,
the average DoF acceleration increases along with $\eta$ when reward terms are kept identical.

\begin{wrapfigure}{r}{0.25\textwidth}
    \vspace{-10pt}
    \centering
    \includegraphics[width=0.25\textwidth]{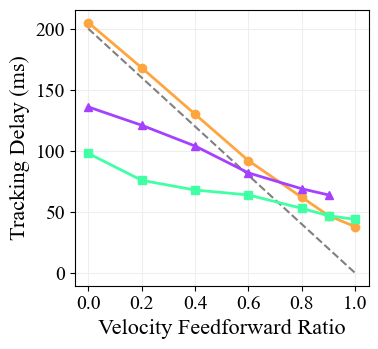}
    \vspace{-15pt}
    \caption{Different tracking delays with respect to velocity feedforward ratios.}
    \label{fig:latency_eta}
    \vspace{-10pt}
\end{wrapfigure}

We illustrate the tracking delay of \orange{velocity feedforward control,} \purple{real-world teleoperation system,} and \cyan{policy rollout in simulation} in Fig.~\ref{fig:latency_eta}, under $\omega_n=10$ rad/s, where the \textcolor{gray}{dashed line} represents the predicted result from Theorem~\ref{thm:one}. Two key observations emerge: (1) the low-level control delay is accurately predicted by Theorem~\ref{thm:one} at low $\eta$, with deviations at high $\eta$ consistent with the discrete-time overshoot described in Theorem~\ref{thm:two}; and (2) the policy is able to compensate for low-level control delay up to $\eta=0.9$, where policy delay matches low-level control delay. Together with the instability observed at $\omega_n=15$ rad/s, these results suggest that {50\,ms represents a practical lower bound on achievable end-to-end teleoperation latency under our current hardware, and that \name~operates at this limit.}

During real-world deployment,
we use PD gains $k_p$ and $k_d$ computed from Eq.~\ref{eq:pd_from_impedance},
with the position target $q^*_t = k \cdot a_t + c$, where $k$ and $c$ are fixed action scaling and offset parameters,
respectively. The velocity target is defined as $\dot{q}^*_t = \eta\cdot(q^*_t - q^*_{t-1})/\Delta t$.
Compared to existing systems, \textbf{the only modification is replacing the zero velocity target with scaled velocity feedforward.}

\orange{\textbf{Q2:} \textbf{How does Direct Extremity Control affect the teleoperation system?}}

The closed-form Cartesian mapping incurs a fixed computational cost of approximately 0.3\,ms regardless of input motion, whereas joint-space retargeting via inverse kinematics introduces variable and potentially unbounded computation time due to the iterative nature of IK solvers.

Benefiting from the one-shot calibration, \orange{Direct Extremity Control requires no human shape prior for retargeting,}
which is particularly advantageous in real-world deployment, because operators may vary significantly in heights, limb lengths, etc.
To evaluate retargeting quality, we compare our method against GMR~\cite{GMR} using five motion capture
sequences collected from two operators, as well as ACCAD dataset, which contains short motion clips from 20 subjects.

\begin{wraptable}{r}{0.35\textwidth}
\vspace{-10pt}
\centering
\setlength{\tabcolsep}{6pt}
\renewcommand{\arraystretch}{1.4}
\setlength{\aboverulesep}{2pt}
\setlength{\belowrulesep}{-1pt}

\scriptsize
\begin{tabular*}{0.32\textwidth}{lcc}
\toprule
\\[-10pt]
\rowcolor{gray!15}
Alignment & GMR & \textbf{Ours} 
\\
\hline
$R_x$ & 0.78$\pm$0.21 & \textbf{0.91$\pm$0.11}
\\
\hline
$R_y$ & 0.62$\pm$0.24 & \textbf{0.76$\pm$0.19}
\\
\hline
$R_z$ & 0.20$\pm$0.26 & \textbf{0.41$\pm$0.15}
\\[-2pt]
\bottomrule
\end{tabular*}

\caption{\textbf{Pearson correlation between hand trajectories before and after retargeting}, averaged across our collected MoCap dataset and ACCAD dataset, weighted by the number of samples in each dataset.}
\vspace{-8pt}
\label{tab:R_retarget}
\end{wraptable}
\orange{Direct Extremity Control achieves higher Pearson correlation in retargeted hand position relative to the original motion}
compared to GMR~\cite{GMR} (Tab.~\ref{tab:R_retarget}), indicating more faithful preservation of the operator's intended motion.
We attribute this to GMR being over-constrained by 14 links, which reduces retargeting flexibility. This is particularly consequential
for manipulation tasks, where end-effector accuracy directly determines whether contacts and grasps are successfully reproduced.
The asymmetry between $x$ and $y$ correlations reflects a directional bias in the MoCap collected dataset, as all sequences were recorded
with the teleoperator facing the x-axis.
The inverse kinematics process after cartesian mapping in GMR introduces 72\,mm and 0.025\,rad error for hands and 24\,mm and 0.044\,rad error for the feet, which
already exceed the policy tracking error as illustrated in Tab.~\ref{tab:tracking_error}. Further more, GMR persists 95.2\% of still-foot frames
defined as foot velocity under 0.25\,m/s, while GMR persists 100\%.

\begin{wraptable}{r}{0.498\textwidth}
\centering
\setlength{\tabcolsep}{6pt}
\renewcommand{\arraystretch}{1.4}
\setlength{\aboverulesep}{2pt}
\setlength{\belowrulesep}{-1pt}
\vspace{-10pt}

\scriptsize
\begin{tabular*}{0.498\textwidth}{lccc}
\toprule
\\[-10pt]
\rowcolor{gray!15}
Observation\hspace{-5pt} &
${\mathbf{T}^r}$ &
$({\mathbf{T}^r},{q^*})$ & 
${q^*}$
\\
\hline
\hspace{-5pt}Tracking Pos Error (mm)\hspace{-5pt} &
49$\pm$7\hspace{-1pt} & \textbf{43$\pm$8}\hspace{-1pt} & 105$\pm$23\hspace{-1pt}
\\
\hline
\hspace{-5pt}Tracking Rot Error (rad) \hspace{-5pt} &
0.19$\pm$0.03\hspace{-1pt} & \textbf{0.17$\pm$0.02}\hspace{-1pt} & 0.48$\pm$0.07\hspace{-1pt}
\\[-2pt]
\bottomrule

\end{tabular*}

\caption{Policy tracking errors in simulation, $\eta=0.9$.}
\vspace{-10pt}
\label{tab:tracking_error}
\end{wraptable}

Although reducing observations inevitably degrades performance, selecting the right observations can substantially mitigate this degradation. Regarding end-effector tracking accuracy, which is critical in loco-manipulation tasks, observing selected link poses alone is inferior to providing both link poses and retargeted joint configurations, yet significantly outperforms observing retargeted joint configurations alone. All policies evaluated in Tab.~\ref{tab:tracking_error} are trained with the same mixed dataset of Cartesian-space mapped motions from our collected dataset with inverse kinematics results and GMR retargeted public motion datasets.

\orange{\textbf{Q3:} \textbf{Does end-to-end latency play a critical role in real-world data collection?}}

\orange{Yes, and this holds consistently across all closed-loop tasks.}
Existing teleoperation systems often introduce delays comparable to the typical human reaction time of approximately 200\,ms, effectively doubling
the response time of the human-in-the-loop system. In contrast, \name achieves an end-to-end latency of only 54\,ms under the optical MoCap setting,
substantially reducing the additional delay imposed on the operator and enabling reaction-sensitive tasks such as the ping-pong return demonstrated
in Fig.~\ref{fig:teaser}.
To evaluate teleoperation performance under varying system-level latencies, we introduce an artificial delay by adding a FIFO queue to the
motion observations received by the policy. We apply delays of 0, 2, 5, and 8 frames, corresponding to approximate overall latencies of 50\,ms, 100\,ms, 150\,ms, and 200\,ms, respectively.

\begin{wraptable}{r}{0.581\textwidth}
\vspace{-10pt}
\centering
\setlength{\tabcolsep}{6pt}
\renewcommand{\arraystretch}{1.4}
\setlength{\aboverulesep}{2pt}
\setlength{\belowrulesep}{-1pt}

\scriptsize
\begin{tabular*}{0.581\textwidth}{lccc}
\toprule
\\[-10pt]
\rowcolor{gray!15}
Latency & Ping-pong Return $\uparrow$ & Ping-pong Balance $\uparrow$ & Cube Stack $\downarrow$
\\
\hline
$\sim50$\,ms & \textbf{80\%} & \textbf{7.3$\pm$3.1\,s} & \textbf{7.6$\pm$2.0\,s}
\\
\hline
$\sim100$\,ms & 60\% & 3.6$\pm$1.7\,s & 8.3$\pm$2.7\,s
\\
\hline
$\sim150$\,ms & 30\% & 1.3$\pm$0.4\,s & 9.5$\pm$2.1\,s
\\
\hline
$\sim200$\,ms & 20\% & 1.5$\pm$0.6\,s & 10.1$\pm3.8$\,s
\\[-2pt]
\bottomrule
\end{tabular*}

\caption{Performance of \name~over 10 trials per task.}
\vspace{-8pt}
\label{tab:real_latency}
\end{wraptable}

As shown in Tab.~\ref{tab:real_latency}, \orange{increasing latency leads to consistent performance degradation across all tasks}. For closed-loop reaction-sensitive tasks, the effect is severe: ping-pong return rate drops from 80\% to 20\% and ball balance duration from 7.3\,s to 1.5\,s as latency increases from 50\,ms to 200\,ms, as delayed visual feedback causes corrective actions to arrive too late. For less time-critical tasks such as cube stacking, the effect is milder but still measurable, with completion time increasing from 7.6\,s to 10.1\,s. These results confirm that human-in-the-loop humanoid control is fundamentally bottlenecked by transmission latency, particularly for tasks demanding rapid closed-loop corrections.
\section{Conclusion and Limitations}


In this work, we present \name, a low-latency humanoid
teleoperation framework combining scaled velocity feedforward
with direct extremity control. Our key findings are:

\begin{itemize}[
    leftmargin=1.2em,
    labelsep=0.5em,
    itemsep=2pt,
    topsep=2pt,
    parsep=0pt,
    partopsep=0pt
]
    \item \
    \name~achieves end-to-end latencies of 54\,ms and 64\,ms
    under MoCap and VR settings, respectively, enabling
    highly dynamic and reaction-sensitive tasks.
    Controlled latency experiments further demonstrate
    the critical importance of low latency for
    closed-loop teleoperation.

    \item 
    Velocity feedforward plays a crucial role in reducing policy tracking delay in teleoperation settings where future motion references are unavailable. It also promotes smoother policy actions by increasing the joint acceleration and torque generated per unit action rate.

    \item 
    Direct extremity control eliminates the need for joint-space retargeting during deployment, avoiding its computational overhead and additional errors introduced by inverse kinematics, at the cost of only a small increase in policy tracking error.
\end{itemize}

Despite these advances, several limitations remain. First, the Unitree G1's seven DoFs per arm introduce inverse kinematics ambiguity, which, combined with direct extremity pose mapping and the distribution of wrist and elbow joints, can lead to unnatural arm poses. Second, while our work primarily targets control-level responsiveness, lower-body latency remains governed by policy-level center-of-mass regulation, such as distinguishing walking initiation from simply lifting a foot. Finally, our framework assumes limited environmental contact and focuses on sparse extremity tracking. Extending it to dense whole-body contact scenarios, such as climbing or crawling, may require additional contact-aware objectives and more sophisticated motion representations.

\acknowledgments{This work was supported by NSF IIS-2441250, NSF IIS-2404386, MURI N000142412748, and Amazon's gift funding. We thank our colleagues for their valuable discussions, insightful feedback, and support throughout this project. The first author would also like to thank his cats, Cookie and Mochi, for their love and companionship.}

\bibliography{references}  

@misc{optitrack,
  author = {NaturalPoint Inc.},
  title = {OptiTrack - Motion Capture Systems},
  year = 2026,
  url = {https://www.optitrack.com/},
  urldate = {2026-01-28}
}

@misc{metaquest,
  author = {Meta.},
  title = {Meta Quest VR Headsets and Accessories | Meta Store},
  year = 2026,
  url = {https://www.meta.com/quest/},
  urldate = {2026-01-28}
}

@misc{vivetracker,
  author = {HTC Corporation},
  title = {VIVE Ultimate Tracker - Full-Body Tracking, SteamVR Support},
  year = 2026,
  url = {https://www.vive.com/us/accessory/vive-ultimate-tracker/},
  urldate = {2026-01-28}
}

@misc{steamvr,
  author = {Valve Corporation},
  title = {SteamVR - Valve Corporation},
  year = 2026,
  url = {https://www.steamvr.com/},
  urldate = {2026-01-28}
}

@misc{beyondmimic,
      title={BeyondMimic: From Motion Tracking to Versatile Humanoid Control via Guided Diffusion}, 
      author={Qiayuan Liao and Takara E. Truong and Xiaoyu Huang and Yuman Gao and Guy Tevet and Koushil Sreenath and C. Karen Liu},
      year={2025},
      eprint={2508.08241},
      archivePrefix={arXiv},
      primaryClass={cs.RO},
      url={https://arxiv.org/abs/2508.08241}, 
}

@article{Hwangbo2019learningagileanddynamic,
author = {Jemin Hwangbo  and Joonho Lee  and Alexey Dosovitskiy  and Dario Bellicoso  and Vassilios Tsounis  and Vladlen Koltun  and Marco Hutter },
title = {Learning agile and dynamic motor skills for legged robots},
journal = {Science Robotics},
volume = {4},
number = {26},
pages = {eaau5872},
year = {2019},
doi = {10.1126/scirobotics.aau5872},
URL = {https://www.science.org/doi/abs/10.1126/scirobotics.aau5872},
eprint = {https://www.science.org/doi/pdf/10.1126/scirobotics.aau5872}}

@article{h20,
      title={Learning human-to-humanoid real-time whole-body teleoperation},
      author={He, Tairan and Luo, Zhengyi and Xiao, Wenli and Zhang, Chong and Kitani, Kris and Liu, Changliu and Shi, Guanya},
      journal={arXiv preprint arXiv:2403.04436},
      year={2024}
    }

@article{omnih2o,
      title={OmniH2O: Universal and Dexterous Human-to-Humanoid Whole-Body Teleoperation and Learning},
      author={He, Tairan and Luo, Zhengyi and He, Xialin and Xiao, Wenli and Zhang, Chong and Zhang, Weinan and Kitani, Kris and Liu, Changliu and Shi, Guanya},
      journal={arXiv preprint arXiv:2406.08858},
      year={2024}
    }

@inproceedings{
      humanoidparkourlearning,
      title={Humanoid Parkour Learning},
      author={Ziwen Zhuang and Shenzhe Yao and Hang Zhao},
      booktitle={8th Annual Conference on Robot Learning},
      year={2024},
      url={https://openreview.net/forum?id=fs7ia3FqUM}
      }

@misc{WoCoCo,
      title={WoCoCo: Learning Whole-Body Humanoid Control with Sequential Contacts}, 
      author={Chong Zhang and Wenli Xiao and Tairan He and Guanya Shi},
      year={2024},
      eprint={2406.06005},
      archivePrefix={arXiv},
      primaryClass={cs.RO}
}

@INPROCEEDINGS{PHC,
  author={Luo, Zhengyi and Cao, Jinkun and Winkler, Alexander and Kitani, Kris and Xu, Weipeng},
  booktitle={2023 IEEE/CVF International Conference on Computer Vision (ICCV)}, 
  title={Perpetual Humanoid Control for Real-time Simulated Avatars}, 
  year={2023},
  volume={},
  number={},
  pages={10861-10870},
  doi={10.1109/ICCV51070.2023.01000}}

@article{AMP,
author = {Peng, Xue Bin and Ma, Ze and Abbeel, Pieter and Levine, Sergey and Kanazawa, Angjoo},
title = {AMP: adversarial motion priors for stylized physics-based character control},
year = {2021},
issue_date = {August 2021},
publisher = {Association for Computing Machinery},
address = {New York, NY, USA},
volume = {40},
number = {4},
issn = {0730-0301},
url = {https://doi.org/10.1145/3450626.3459670},
doi = {10.1145/3450626.3459670},
journal = {ACM Trans. Graph.},
month = jul,
articleno = {144},
numpages = {20}
}

@INPROCEEDINGS{HOVER,
  author={He, Tairan and Xiao, Wenli and Lin, Toru and Luo, Zhengyi and Xu, Zhenjia and Jiang, Zhenyu and Kautz, Jan and Liu, Changliu and Shi, Guanya and Wang, Xiaolong and Fan, Linxi Jim and Zhu, Yuke},
  booktitle={2025 IEEE International Conference on Robotics and Automation (ICRA)}, 
  title={HOVER: Versatile Neural Whole-Body Controller for Humanoid Robots}, 
  year={2025},
  volume={},
  number={},
  pages={9989-9996},
  doi={10.1109/ICRA55743.2025.11128549}}

@inproceedings{HumanPlus,
  author    = {Fu, Zipeng and Zhao, Qingqing and Wu, Qi and Wetzstein, Gordon and Finn, Chelsea},
  title     = {HumanPlus: Humanoid Shadowing and Imitation from Humans},
  booktitle = {{Conference on Robot Learning (CoRL)}},
  year      = {2024},
}

@INPROCEEDINGS{ASAP, 
    AUTHOR    = {Tairan He AND Jiawei Gao AND Wenli Xiao AND Yuanhang Zhang AND Zi Wang AND Jiashun Wang AND Zhengyi Luo AND Guanqi He AND Nikhil Sobanbabu AND Chaoyi Pan AND Zeji Yi AND Guannan Qu AND Kris Kitani AND Jessica K. Hodgins AND Linxi Fan AND Yuke Zhu AND Changliu Liu AND Guanya Shi}, 
    TITLE     = {{ASAP: Aligning Simulation and Real-World Physics for Learning Agile Humanoid Whole-Body Skills}}, 
    BOOKTITLE = {Proceedings of Robotics: Science and Systems}, 
    YEAR      = {2025}, 
    ADDRESS   = {LosAngeles, CA, USA}, 
    MONTH     = {June}, 
    DOI       = {10.15607/RSS.2025.XXI.066} 
}

@misc{GMT,
      title={GMT: General Motion Tracking for Humanoid Whole-Body Control}, 
      author={Zixuan Chen and Mazeyu Ji and Xuxin Cheng and Xuanbin Peng and Xue Bin Peng and Xiaolong Wang},
      year={2025},
      eprint={2506.14770},
      archivePrefix={arXiv},
      primaryClass={cs.RO},
      url={https://arxiv.org/abs/2506.14770}, 
}

@inproceedings{videomimic,
  title     = {Visual imitation enables contextual humanoid control},
  author    = {Allshire, Arthur and Choi, Hongsuk and Zhang, Junyi and McAllister, David 
               and Zhang, Anthony and Kim, Chung Min and Darrell, Trevor and Abbeel, 
               Pieter and Malik, Jitendra and Kanazawa, Angjoo},
  booktitle = {Proceedings of the Conference on Robot Learning (CoRL)},
  year      = {2025}
}

@inproceedings{
TWIST,
title={{TWIST}: Teleoperated Whole-Body Imitation System},
author={Yanjie Ze and Zixuan Chen and Joao Pedro Araujo and Zi-ang Cao and Xue Bin Peng and Jiajun Wu and Karen Liu},
booktitle={9th Annual Conference on Robot Learning},
year={2025},
url={https://openreview.net/forum?id=htgNQHa6Ta}
}

@misc{TWIST2,
      title={TWIST2: Scalable, Portable, and Holistic Humanoid Data Collection System}, 
      author={Yanjie Ze and Siheng Zhao and Weizhuo Wang and Angjoo Kanazawa and Rocky Duan and Pieter Abbeel and Guanya Shi and Jiajun Wu and C. Karen Liu},
      year={2025},
      eprint={2511.02832},
      archivePrefix={arXiv},
      primaryClass={cs.RO},
      url={https://arxiv.org/abs/2511.02832}, 
}

@misc{AMS,
      title={Agility Meets Stability: Versatile Humanoid Control with Heterogeneous Data}, 
      author={Yixuan Pan and Ruoyi Qiao and Li Chen and Kashyap Chitta and Liang Pan and Haoguang Mai and Qingwen Bu and Hao Zhao and Cunyuan Zheng and Ping Luo and Hongyang Li},
      year={2025},
      eprint={2511.17373},
      archivePrefix={arXiv},
      primaryClass={cs.RO},
      url={https://arxiv.org/abs/2511.17373}, 
}

@INPROCEEDINGS{AMO, 
    AUTHOR    = {Jialong Li AND Xuxin Cheng AND Tianshu Huang AND Shiqi Yang AND Ri-Zhao Qiu AND Xiaolong Wang}, 
    TITLE     = {{AMO: Adaptive Motion Optimization for Hyper-Dexterous Humanoid Whole-Body Control}}, 
    BOOKTITLE = {Proceedings of Robotics: Science and Systems}, 
    YEAR      = {2025}, 
    ADDRESS   = {LosAngeles, CA, USA}, 
    MONTH     = {June}, 
    DOI       = {10.15607/RSS.2025.XXI.061} 
}

@misc{GMR,
      title={Retargeting Matters: General Motion Retargeting for Humanoid Motion Tracking}, 
      author={Joao Pedro Araujo and Yanjie Ze and Pei Xu and Jiajun Wu and C. Karen Liu},
      year={2025},
      eprint={2510.02252},
      archivePrefix={arXiv},
      primaryClass={cs.RO},
      url={https://arxiv.org/abs/2510.02252}, 
}

@misc{PPO,
      title={Proximal Policy Optimization Algorithms}, 
      author={John Schulman and Filip Wolski and Prafulla Dhariwal and Alec Radford and Oleg Klimov},
      year={2017},
      eprint={1707.06347},
      archivePrefix={arXiv},
      primaryClass={cs.LG},
      url={https://arxiv.org/abs/1707.06347}, 
}

@InProceedings{DAGGer,
  title = 	 {A Reduction of Imitation Learning and Structured Prediction to No-Regret Online Learning},
  author = 	 {Ross, Stephane and Gordon, Geoffrey and Bagnell, Drew},
  booktitle = 	 {Proceedings of the Fourteenth International Conference on Artificial Intelligence and Statistics},
  pages = 	 {627--635},
  year = 	 {2011},
  editor = 	 {Gordon, Geoffrey and Dunson, David and Dudík, Miroslav},
  volume = 	 {15},
  series = 	 {Proceedings of Machine Learning Research},
  address = 	 {Fort Lauderdale, FL, USA},
  month = 	 {11--13 Apr},
  publisher =    {PMLR},
  url = 	 {https://proceedings.mlr.press/v15/ross11a.html}
}

@misc{Genesis,
          author = {Genesis Authors},
          title = {Genesis: A Generative and Universal Physics Engine for Robotics and Beyond},
          month = {December},
          year = {2024},
          url = {https://github.com/Genesis-Embodied-AI/Genesis}
        }

@inproceedings{farneback2003two,
  title={Two-frame motion estimation based on polynomial expansion},
  author={Farneb{\"a}ck, Gunnar},
  booktitle={Scandinavian conference on Image analysis},
  pages={363--370},
  year={2003},
  organization={Springer}
}

@INPROCEEDINGS{ExpressiveWBCHumanoid, 
    AUTHOR    = {Xuxin Cheng AND Yandong Ji AND Junming Chen AND Ruihan Yang AND Ge Yang AND Xiaolong Wang}, 
    TITLE     = {{Expressive Whole-Body Control for Humanoid Robots}}, 
    BOOKTITLE = {Proceedings of Robotics: Science and Systems}, 
    YEAR      = {2024}, 
    ADDRESS   = {Delft, Netherlands}, 
    MONTH     = {July}, 
    DOI       = {10.15607/RSS.2024.XX.107} 
}

@INPROCEEDINGS{Sim-to-Real, 
    AUTHOR    = {Jie Tan AND Tingnan Zhang AND Erwin Coumans AND Atil Iscen AND Yunfei Bai AND Danijar Hafner AND Steven Bohez AND Vincent Vanhoucke}, 
    TITLE     = {Sim-to-Real: Learning Agile Locomotion For Quadruped Robots}, 
    BOOKTITLE = {Proceedings of Robotics: Science and Systems}, 
    YEAR      = {2018}, 
    ADDRESS   = {Pittsburgh, Pennsylvania}, 
    MONTH     = {June}, 
    DOI       = {10.15607/RSS.2018.XIV.010} 
}

@inproceedings{
LearningToWalk,
title={Learning to Walk in Minutes Using Massively Parallel Deep Reinforcement Learning},
author={Nikita Rudin and David Hoeller and Philipp Reist and Marco Hutter},
booktitle={5th Annual Conference on Robot Learning },
year={2021},
url={https://openreview.net/forum?id=wK2fDDJ5VcF}
}

@article{ImitatingAndFinetuningQuadrupedal,
      title={Imitating and Finetuning Model Predictive Control for Robust and Symmetric Quadrupedal Locomotion},
      author={Youm, Donghoon and Jung, Hyunyoung and Kim, Hyeongjun and Hwangbo, Jemin and Park, Hae-Won and Ha, Sehoon},
      journal={IEEE Robotics and Automation Letters},
      year={2023},
      publisher={IEEE}
    }

@InProceedings{ContinuousVersatileJumping,
  title = 	 {Continuous Versatile Jumping Using Learned Action Residuals},
  author =       {Yang, Yuxiang and Meng, Xiangyun and Yu, Wenhao and Zhang, Tingnan and Tan, Jie and Boots, Byron},
  booktitle = 	 {Proceedings of The 5th Annual Learning for Dynamics and Control Conference},
  pages = 	 {770--782},
  year = 	 {2023},
  editor = 	 {Matni, Nikolai and Morari, Manfred and Pappas, George J.},
  volume = 	 {211},
  series = 	 {Proceedings of Machine Learning Research},
  month = 	 {15--16 Jun},
  publisher =    {PMLR},
  url = 	 {https://proceedings.mlr.press/v211/yang23b.html}
}

@INPROCEEDINGS{ResidualRLRobotControl,
  author={Johannink, Tobias and Bahl, Shikhar and Nair, Ashvin and Luo, Jianlan and Kumar, Avinash and Loskyll, Matthias and Ojea, Juan Aparicio and Solowjow, Eugen and Levine, Sergey},
  booktitle={2019 International Conference on Robotics and Automation (ICRA)}, 
  title={Residual Reinforcement Learning for Robot Control}, 
  year={2019},
  volume={},
  number={},
  pages={6023-6029},
  doi={10.1109/ICRA.2019.8794127}}

@article{SYNTHESISWBC,
author = {SENTIS, LUIS and KHATIB, OUSSAMA},
title = {SYNTHESIS OF WHOLE-BODY BEHAVIORS THROUGH HIERARCHICAL CONTROL OF BEHAVIORAL PRIMITIVES},
journal = {International Journal of Humanoid Robotics},
volume = {02},
number = {04},
pages = {505-518},
year = {2005},
doi = {10.1142/S0219843605000594},

URL = { https://doi.org/10.1142/S0219843605000594
},
eprint = { 
https://doi.org/10.1142/S0219843605000594
}
}

@INPROCEEDINGS{HumanoidWBCFramework,
  author={Sentis, L. and Khatib, O.},
  booktitle={Proceedings 2006 IEEE International Conference on Robotics and Automation, 2006. ICRA 2006.}, 
  title={A whole-body control framework for humanoids operating in human environments}, 
  year={2006},
  volume={},
  number={},
  pages={2641-2648},
  doi={10.1109/ROBOT.2006.1642100}}

@article{TorqueControllHumanoid,
  title={Balancing experiments on a torque-controlled humanoid with hierarchical inverse dynamics},
  author={Alexander Herzog and Ludovic Righetti and Felix Grimminger and Peter Pastor and Stefan Schaal},
  journal={2014 IEEE/RSJ International Conference on Intelligent Robots and Systems},
  year={2013},
  pages={981-988},
  url={https://api.semanticscholar.org/CorpusID:3104000}
}

@INPROCEEDINGS{AtlasFullBodyControl,
  author={Feng, Siyuan and Whitman, Eric and Xinjilefu, X and Atkeson, Christopher G.},
  booktitle={2014 IEEE-RAS International Conference on Humanoid Robots}, 
  title={Optimization based full body control for the atlas robot}, 
  year={2014},
  volume={},
  number={},
  pages={120-127},
  doi={10.1109/HUMANOIDS.2014.7041347}}

@article{PassiveBasedTorqueControl,
author = {Bernd Henze and Máximo A. Roa and Christian Ott},
title ={Passivity-based whole-body balancing for torque-controlled humanoid robots in
          multi-contact scenarios},

journal = {The International Journal of Robotics Research},
volume = {35},
number = {12},
pages = {1522-1543},
year = {2016},
doi = {10.1177/0278364916653815},

URL = { 
    
        https://doi.org/10.1177/0278364916653815
    
    

},
eprint = { 
    
        https://doi.org/10.1177/0278364916653815
    
    

}
}

@INPROCEEDINGS{MobileTeleVision,
  author={Lu, Chenhao and Cheng, Xuxin and Li, Jialong and Yang, Shiqi and Ji, Mazeyu and Yuan, Chengjing and Yang, Ge and Yi, Sha and Wang, Xiaolong},
  booktitle={2025 IEEE International Conference on Robotics and Automation (ICRA)}, 
  title={Mobile-TeleVision: Predictive Motion Priors for Humanoid Whole-Body Control}, 
  year={2025},
  volume={},
  number={},
  pages={5364-5371},
  doi={10.1109/ICRA55743.2025.11128652}}

@INPROCEEDINGS{HumanoidPose,
      author={Mao, Jiageng and Zhao, Siheng and Song, Siqi and Hong, Chuye and Shi, Tianheng and Ye, Junjie and Zhang, Mingtong and Geng, Haoran and Malik, Jitendra and Guizilini, Vitor and Wang, Yue},
      booktitle={2025 IEEE-RAS 24th International Conference on Humanoid Robots (Humanoids)}, 
      title={Universal Humanoid Robot Pose Learning from Internet Human Videos}, 
      year={2025},
      volume={},
      number={},
      pages={1-8},
      doi={10.1109/Humanoids65713.2025.11203143}}

@misc{HiFAR,
      title={HiFAR: Multi-Stage Curriculum Learning for High-Dynamics Humanoid Fall Recovery}, 
      author={Penghui Chen and Yushi Wang and Changsheng Luo and Wenhan Cai and Mingguo Zhao},
      year={2025},
      eprint={2502.20061},
      archivePrefix={arXiv},
      primaryClass={cs.RO},
      url={https://arxiv.org/abs/2502.20061}, 
}

@misc{HoldMyBeer,
      title={Hold My Beer: Learning Gentle Humanoid Locomotion and End-Effector Stabilization Control}, 
      author={Yitang Li and Yuanhang Zhang and Wenli Xiao and Chaoyi Pan and Haoyang Weng and Guanqi He and Tairan He and Guanya Shi},
      year={2025},
      eprint={2505.24198},
      archivePrefix={arXiv},
      primaryClass={cs.RO},
      url={https://arxiv.org/abs/2505.24198}, 
}

@inproceedings{mahmood2019amass,
  title={AMASS: Archive of motion capture as surface shapes},
  author={Mahmood, Naureen and Ghorbani, Nima and Troje, Nikolaus F and Pons-Moll, Gerard and Black, Michael J},
  booktitle={Proceedings of the IEEE/CVF international conference on computer vision},
  pages={5442--5451},
  year={2019}
}

@article{harvey2020robust,
  title={Robust motion in-betweening},
  author={Harvey, F{\'e}lix G and Yurick, Mike and Nowrouzezahrai, Derek and Pal, Christopher},
  journal={ACM Transactions on Graphics (TOG)},
  volume={39},
  number={4},
  pages={60--1},
  year={2020},
  publisher={ACM New York, NY, USA}
}

@misc{lv2025lafan1_retargeting,
  author       = {H. Lv},
  title        = {LAFAN1 Retargeting Dataset},
  year         = {2025},
  howpublished = {\url{https://huggingface.co/datasets/lvhaidong/LAFAN1_Retargeting_Dataset}},
}

@misc{omniretarget,
      title={OmniRetarget: Interaction-Preserving Data Generation for Humanoid Whole-Body Loco-Manipulation and Scene Interaction}, 
      author={Lujie Yang and Xiaoyu Huang and Zhen Wu and Angjoo Kanazawa and Pieter Abbeel and Carmelo Sferrazza and C. Karen Liu and Rocky Duan and Guanya Shi},
      year={2025},
      eprint={2509.26633},
      archivePrefix={arXiv},
      primaryClass={cs.RO},
      url={https://arxiv.org/abs/2509.26633}, 
}

@article{Hassaan2014TuningOA,
  title={Tuning of a PD controller used with second order processes},
  author={Galal A. Hassaan},
  journal={International Journal of Engineering},
  year={2014},
  volume={2},
  url={https://api.semanticscholar.org/CorpusID:195955908}
}

@InProceedings{leggedgym,
  title = 	 {Learning to Walk in Minutes Using Massively Parallel Deep Reinforcement Learning},
  author =       {Rudin, Nikita and Hoeller, David and Reist, Philipp and Hutter, Marco},
  booktitle = 	 {Proceedings of the 5th Conference on Robot Learning},
  pages = 	 {91--100},
  year = 	 {2022},
  editor = 	 {Faust, Aleksandra and Hsu, David and Neumann, Gerhard},
  volume = 	 {164},
  series = 	 {Proceedings of Machine Learning Research},
  month = 	 {08--11 Nov},
  publisher =    {PMLR},
  url = 	 {https://proceedings.mlr.press/v164/rudin22a.html}
}

@inproceedings{
HuB,
title={HuB: Learning Extreme Humanoid Balance},
author={Tong Zhang and Boyuan Zheng and Ruiqian Nai and Yingdong Hu and Yen-Jen Wang and Geng Chen and Fanqi Lin and Jiongye Li and Chuye Hong and Koushil Sreenath and Yang Gao},
booktitle={9th Annual Conference on Robot Learning},
year={2025},
url={https://openreview.net/forum?id=FCpYuGtN4j}
}

@inproceedings{SPARK,
              title        = {{SPARK}: Safe Protective and Assistive Robot Kit},
              author       = {Sun, Yifan and
                              Chen, Rui and
                              Yun, Kai~S. and
                              Fang, Yikuan and
                              Jung, Sebin and
                              Li, Feihan and
                              Li, Bowei and
                              Zhao, Weiye and
                              Liu, Changliu},
              booktitle    = {IFAC Symposium on Robotics},
              year         = {2025},
              eprint       = {2502.03132},
              archivePrefix= {arXiv},
              primaryClass = {cs.RO},
              url          = {https://intelligent-control-lab.github.io/spark/},
              video        = {https://www.youtube.com/embed/vIzeQ31YbCM},
              repository   = {https://github.com/intelligent-control-lab/spark}
            }

@article{KungfuBot2,
      title={KungfuBot2: Learning Versatile Motion Skills for Humanoid Whole-Body Control}, 
      author={Han, Jinrui and Xie, Weiji and Zheng, Jiakun and Shi, Jiyuan and Zhang, Weinan and Xiao, Ting and Bai, Chenjia},
      journal={arXiv:2509.16638},
      year={2025}
    }

@misc{SONIC,
      title={SONIC: Supersizing Motion Tracking for Natural Humanoid Whole-Body Control}, 
      author={Zhengyi Luo and Ye Yuan and Tingwu Wang and Chenran Li and Sirui Chen and Fernando Castañeda and Zi-Ang Cao and Jiefeng Li and David Minor and Qingwei Ben and Xingye Da and Runyu Ding and Cyrus Hogg and Lina Song and Edy Lim and Eugene Jeong and Tairan He and Haoru Xue and Wenli Xiao and Zi Wang and Simon Yuen and Jan Kautz and Yan Chang and Umar Iqbal and Linxi "Jim" Fan and Yuke Zhu},
      year={2025},
      eprint={2511.07820},
      archivePrefix={arXiv},
      primaryClass={cs.RO},
      url={https://arxiv.org/abs/2511.07820}, 
}

@INPROCEEDINGS{HeavyLifting,
  author={Purushottam, Amartya and Yan, Jack and Xu, Christopher and Ramos, Joao},
  booktitle={2025 IEEE-RAS 24th International Conference on Humanoid Robots (Humanoids)}, 
  title={Heavy Lifting Tasks via Haptic Teleoperation of a Wheeled Humanoid}, 
  year={2025},
  volume={},
  number={},
  pages={345-350},
  doi={10.1109/Humanoids65713.2025.11203084}}

@InProceedings{CLONE,
  title = 	 {CLONE: Closed-Loop Whole-Body Humanoid Teleoperation for Long-Horizon Tasks},
  author =       {Li, Yixuan and Lin, Yutang and Cui, Jieming and Liu, Tengyu and Liang, Wei and Zhu, Yixin and Huang, Siyuan},
  booktitle = 	 {Proceedings of The 9th Conference on Robot Learning},
  pages = 	 {4493--4505},
  year = 	 {2025},
  editor = 	 {Lim, Joseph and Song, Shuran and Park, Hae-Won},
  volume = 	 {305},
  series = 	 {Proceedings of Machine Learning Research},
  month = 	 {27--30 Sep},
  publisher =    {PMLR},
  url = 	 {https://proceedings.mlr.press/v305/li25h.html}
}

@INPROCEEDINGS{HOMIE, 
    AUTHOR    = {Qingwei Ben AND Feiyu Jia AND Jia Zeng AND Junting Dong AND Dahua Lin AND Jiangmiao Pang}, 
    TITLE     = {{HOMIE: Humanoid Loco-Manipulation with Isomorphic Exoskeleton Cockpit}}, 
    BOOKTITLE = {Proceedings of Robotics: Science and Systems}, 
    YEAR      = {2025}, 
    ADDRESS   = {LosAngeles, CA, USA}, 
    MONTH     = {June}, 
    DOI       = {10.15607/RSS.2025.XXI.070} 
}

@INPROCEEDINGS{CHILD,
  author={Myers, Noboru and Kwon, Obin and Yamsani, Sankalp and Kim, Joohyung},
  booktitle={2025 IEEE-RAS 24th International Conference on Humanoid Robots (Humanoids)}, 
  title={CHILD (Controller for Humanoid Imitation and Live Demonstration): A Whole-Body Humanoid Teleoperation System}, 
  year={2025},
  volume={},
  number={},
  pages={1-6},
  doi={10.1109/Humanoids65713.2025.11203119}}

@inproceedings{
ACE,
title={{ACE}: A Cross-platform and visual-Exoskeletons System for Low-Cost Dexterous Teleoperation},
author={Shiqi Yang and Minghuan Liu and Yuzhe Qin and Runyu Ding and Jialong Li and Xuxin Cheng and Ruihan Yang and Sha Yi and Xiaolong Wang},
booktitle={8th Annual Conference on Robot Learning},
year={2024},
url={https://openreview.net/forum?id=7ddT4eklmQ}
}

@inproceedings{RMA,
title={Rma: Rapid motor adaptation for legged robots},
author={Kumar, Ashish and Fu, Zipeng and Pathak, Deepak and Malik, Jitendra},
journal={Robotics: Science and Systems},
year={2021}
}

@inproceedings{
LearningRobustRewards,
title={Learning Robust Rewards with Adverserial Inverse Reinforcement Learning},
author={Justin Fu and Katie Luo and Sergey Levine},
booktitle={International Conference on Learning Representations},
year={2018},
url={https://openreview.net/forum?id=rkHywl-A-},
}

@inproceedings{robotparkourlearning,
  author    = {Zhuang, Ziwen and Fu, Zipeng and Wang, Jianren and Atkeson, Christopher and Schwertfeger, Sören and Finn, Chelsea and Zhao, Hang},
  title     = {Robot Parkour Learning},
  booktitle = {Conference on Robot Learning ({CoRL})},
  year      = {2023},
}

@InProceedings{walktheseways,
  title = 	 {Walk These Ways: Tuning Robot Control for Generalization with Multiplicity of Behavior},
  author =       {Margolis, Gabriel B. and Agrawal, Pulkit},
  booktitle = 	 {Proceedings of The 6th Conference on Robot Learning},
  pages = 	 {22--31},
  year = 	 {2023},
  editor = 	 {Liu, Karen and Kulic, Dana and Ichnowski, Jeff},
  volume = 	 {205},
  series = 	 {Proceedings of Machine Learning Research},
  month = 	 {14--18 Dec},
  publisher =    {PMLR},
  url = 	 {https://proceedings.mlr.press/v205/margolis23a.html}
}

@misc{bfmzero,
      title={BFM-Zero: A Promptable Behavioral Foundation Model for Humanoid Control Using Unsupervised Reinforcement Learning}, 
      author={Yitang Li and Zhengyi Luo and Tonghe Zhang and Cunxi Dai and Anssi Kanervisto and Andrea Tirinzoni and Haoyang Weng and Kris Kitani and Mateusz Guzek and Ahmed Touati and Alessandro Lazaric and Matteo Pirotta and Guanya Shi},
      year={2025},
      eprint={2511.04131},
      archivePrefix={arXiv},
      primaryClass={cs.RO},
      url={https://arxiv.org/abs/2511.04131}, 
}

@inproceedings{
	2017-SCA-action,
	author = {Peng, Xue Bin and van de Panne, Michiel},
	title = {Learning Locomotion Skills Using DeepRL: Does the Choice of Action Space Matter?},
	booktitle = {Proceedings of the ACM SIGGRAPH / Eurographics Symposium on Computer Animation},
	series = {SCA '17},
	year = {2017},
	isbn = {978-1-4503-5091-4},
	location = {Los Angeles, California},
	pages = {12:1--12:13},
	articleno = {12},
	numpages = {13},
	url = {http://doi.acm.org/10.1145/3099564.3099567},
	doi = {10.1145/3099564.3099567},
	acmid = {3099567},
	publisher = {ACM},
	address = {New York, NY, USA},
}

@misc{xiong2024mqe,
      title={MQE: Unleashing the Power of Interaction with Multi-agent Quadruped Environment}, 
      author={Ziyan Xiong and Bo Chen and Shiyu Huang and Wei-Wei Tu and Zhaofeng He and Yang Gao},
      year={2024},
      eprint={2403.16015},
      archivePrefix={arXiv},
      primaryClass={cs.RO}
}

@misc{fang2026spotspatialperceptionorientedlonghorizon,
      title={SPOT: Spatial Perception-Oriented Long-Horizon Humanoid Teleoperation}, 
      author={Lixing Fang and Ziyan Xiong and Sunli Chen and Zhiyang Dou and Chuang Gan},
      year={2026},
      eprint={2609.07933},
      archivePrefix={arXiv},
      primaryClass={cs.RO},
      url={https://arxiv.org/abs/2609.07933}, 
}

\appendix
\newpage
\section{Additional Proof}
\label{sec:proof}

\subsection{Whole-Body Impedance Calibration}
\label{sec:proof_impedance}
Define the displacement $x \triangleq q - q_0$. Substituting into the governing equation yields
\begin{equation}
    M_{\mathrm{eff}}\,\ddot{x} + k_p\, x = 0,
\end{equation}
which is the equation of an undamped simple harmonic oscillator (the derivative gain having been set to zero by construction). Seeking a solution of the form $x(t) = A e^{i\omega t}$ and substituting gives the characteristic equation
\begin{equation}
    -M_{\mathrm{eff}}\,\omega^2 + k_p = 0
    \implies
    \omega = \sqrt{\frac{k_p}{M_{\mathrm{eff}}}}.
\end{equation}
Since the oscillation period satisfies $P = 2\pi/\omega$, it follows that
\begin{equation}
    P = 2\pi\sqrt{\frac{M_{\mathrm{eff}}}{k_p}}.
\end{equation}
Squaring both sides and solving for $M_{\mathrm{eff}}$ yields
\begin{equation}
    M_{\mathrm{eff}} = \frac{k_p\, P^2}{(2\pi)^2}.
\end{equation}
To obtain an estimate that is robust to modeling errors and sensor noise, we sample $k_p^{(e)} \sim \mathcal{U}(0.5\,k_p, 1.5\,k_p)$ independently across parallel environments $e \in E$. Since the relation above holds for any positive gain value, applying it environment-wise gives
\begin{equation}
    M_{\mathrm{eff}}^{(e)} = \frac{k_p^{(e)}\left(P^{(e)}\right)^2}{(2\pi)^2},
\end{equation}
where $P^{(e)}$ denotes the oscillation period measured in environment $e$. The final estimate is taken as the ensemble mean
\begin{equation}
    \overline{M_{\mathrm{eff}}} = \frac{1}{|E|}\sum_{e \in E} M_{\mathrm{eff}}^{(e)},
\end{equation}
whose variance decreases as $\mathcal{O}(|E|^{-1})$ by the law of large numbers, yielding a statistically consistent estimator.

\subsection{Theoretical Tracking Delay with Scaled Velocity Feedforward}
\label{sec:proof_delay}

For analysis, we model the joint dynamics as $M_{\mathrm{eff}}\ddot{q} = \tau$.
Following the whole-body impedance calibration described in Sec.~\ref{sec:ffcontrol},
we parameterize the PD gains using a target natural frequency $\omega_n$ with critical damping $\zeta = 1$.
Substituting into the closed-loop dynamics yields
\begin{equation}
    \ddot{q} + 2\omega_n \dot{q} + \omega_n^2 q
    = 2\eta\,\omega_n \dot{q}_t + \omega_n^2 q_t.
\end{equation}
Since the closed-loop system is linear and time-invariant, its response to an arbitrary
reference trajectory can be expressed as a superposition of responses to sinusoidal
components via Fourier decomposition. Therefore, without loss of generality, we analyze
a sinusoidal reference
\begin{equation}
    q_t(t) = A\sin(\omega t),
\end{equation}
which fully characterizes the frequency-dependent tracking behavior of the controller.
Taking the Laplace transform, the transfer function from reference $q_t$ to realized
joint position $q$ is
\begin{equation}
    H(s) = \frac{\omega_n^2 + 2\eta\,\omega_n s}{s^2 + 2\omega_n s + \omega_n^2}.
\end{equation}
For a sinusoidal input, the steady-state response takes the form
\begin{equation}
    q(t) = A\,|H(j\omega)|\sin\!\big(\omega t + \phi(\omega)\big),
\end{equation}
where the phase lead between the realized motion and the reference is
\begin{equation}
    \phi(\omega)
    = \tan^{-1}\!\left(\frac{2\eta\,\omega}{\omega_n}\right)
    - \tan^{-1}\!\left(\frac{2(\omega/\omega_n)}{1 - (\omega/\omega_n)^2}\right).
\end{equation}
In the low-frequency regime $\omega \ll \omega_n$, the phase shift admits an interpretation
as an equivalent tracking delay,
\begin{equation}
    \ell = -\frac{\phi(\omega)}{\omega} \approx \frac{2(1-\eta)}{\omega_n}.
    \label{eq:response_latency}
\end{equation}

\subsection{Velocity Feedforward Ratio Upper Bound}
\label{sec:proof_uppperbound}
We derive a practical upper bound on the feedforward ratio $\eta$ by analyzing the
torque balance within a single control interval of duration $\Delta t$.
We assume that at the beginning of each control interval the joint position matches
the reference, $q \approx q_t$, and that the desired trajectory evolves linearly
within the interval at constant velocity $\dot{q}_t$.

Under zero-order hold, the target position is fixed at the value commanded at the
start of the interval, while the true reference advances continuously. The average
positional deviation accumulated between the held target and the true reference
over one control step is therefore
\begin{equation}
    \overline{\Delta q_t} \approx \frac{\dot{q}_t \,\Delta t}{2}.
\end{equation}
This deviation gives rise to two competing torque contributions: a proportional
restoring torque driven by the position error, and a velocity feedforward torque
injected to anticipate the reference motion,
\begin{equation}
    \tau_p \approx k_p \frac{\dot{q}_t\,\Delta t}{2},
    \qquad
    \tau_v \approx \eta\, k_d\, \dot{q}_t.
\end{equation}
To prevent these combined contributions from inducing additional acceleration
within the interval that could cause overshoot, we require their sum to remain
within the damping budget $k_d \dot{q}_t$, i.e.,
\begin{equation}
    \tau_p + \tau_v \leq k_d\, \dot{q}_t.
\end{equation}
Substituting $\tau_p$ and $\tau_v$, and using the whole-body impedance calibration results described in Sec.~\ref{sec:ffcontrol}, the inequality simplifies to
\begin{equation}
    \frac{\omega_n \Delta t}{4} + \eta \leq 1,
\end{equation}
which yields the following upper bound on the admissible feedforward ratio,
\begin{equation}
    \eta \leq 1 - \frac{\omega_n\,\Delta t}{4}.
    \label{eq:eta_bound}
\end{equation}
Equation~\eqref{eq:eta_bound} reveals a fundamental trade-off: higher natural
frequencies $\omega_n$ or longer control intervals $\Delta t$ necessitate a more
conservative feedforward ratio to maintain stable, non-overshooting tracking.

\newpage
\section{Additional Evaluation}

\label{sec:additional_evaluation}

\subsection{Simulation Setup}
\label{sec:simulation_setup}

We leverage Genesis \cite{Genesis} as the simulation backbone, with a simulation timestep of $\mathrm{sim\_dt} = \tfrac{1}{200}\,\mathrm{s}$ and a single physics substep across all experiments, including whole-body impedance calibration, velocity feedforward ratio validation, and policy training. During policy training, we run 8,192 parallel environments, achieving over 100k policy steps per second on an NVIDIA L40s GPU with a control decimation of 4. Detailed simulation parameters and domain randomizations are provided in the appendix.

\subsection{Real-World Robot Setup}
\label{sec:real_world_setup}

Following common practice in control literature \cite{Hassaan2014TuningOA}, we set the PD gains in simulation based on a target natural frequency of $\omega_n = 10\,\mathrm{rad}/\mathrm{s}$ with a damping ratio $\zeta = 1$. The velocity feedforward ratio is set to 0.9, while it is disabled for all indirect-drive joints, which are ankle and waist joints for Unitree G1. The low-level PD controller operates at 1000\,Hz, whereas the high-level control policy runs at 50\,Hz. This setting yields an upper bound on the velocity feedforward ratio of 0.95. To ensure consistent forward kinematics in observation computation, we apply a low-pass filter with coefficient $\alpha = 0.1$ at 1000\,Hz to smooth the measured joint configuration.

\subsection{MoCap System Setup}

We employ an optical motion capture system, OptiTrack~\cite{optitrack}, to obtain the global position and rotation of human body links. Although the OptiTrack Motive software is capable of reconstructing a full-body human pose, our system only utilizes the subset of links described in Sec.~\ref{sec:method_retarget}, which are sufficient for the proposed Cartesian-space mapping and tracking framework.

Human motion is streamed at $120$\,Hz. In contrast to TWIST \cite{TWIST}, we implement a non-blocking streaming pipeline specifically designed for real-time control rather than offline recording. The pipeline minimizes transmission latency such that, when link poses are queried, the effective delay is bounded within a single MoCap streaming cycle ($\frac{1}{120}$\,s). In a local setup, the system latency is less than 10\,ms.

\subsection{VR System Setup}

We employ a VR-based input system composed of a Meta Quest~2~\cite{metaquest} headset, hand controllers, and three VIVE Ultimate Trackers~\cite{vivetracker}, with SteamVR~\cite{steamvr} used as a unifying middleware to resolve cross-device compatibility. The trackers are mounted on the waist and both feet as in Fig.~\ref{fig:calibration}, providing direct pose measurements for the corresponding body links.

Because the torso pose is not directly tracked, we estimate the torso orientation from the planar motion of the headset expressed in the pelvis frame. Specifically, we compute the rotation that aligns the vertical axis $\vec{z}$ with $(\mathbf{R}_{\text{pelvis}})^{-1}\mathbf{p}_{\text{headset}}$, along the rotation axis orthogonal to the plane spanned by them. The headset’s own orientation is intentionally not used, allowing the user to freely observe the environment.

\subsection{System-Level Latency Breakdown}
\label{sec:latency_breakdown}

\begin{figure}[h]
    \centering
    \includegraphics[width=0.9\textwidth]{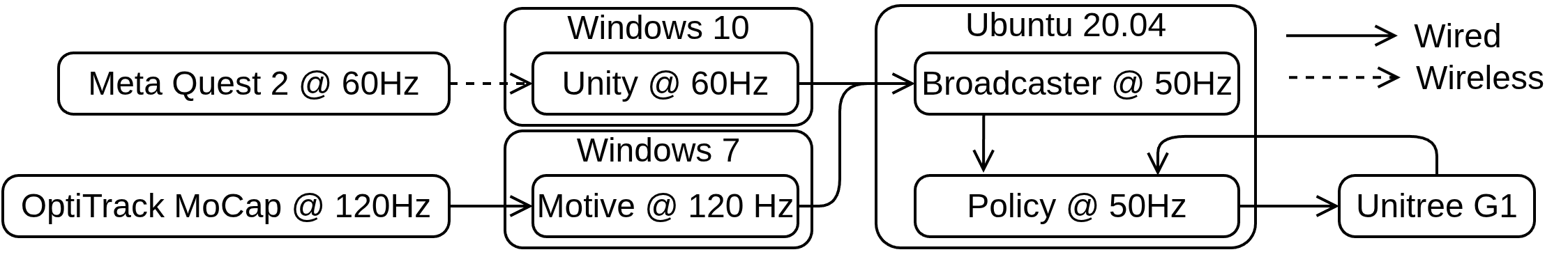}
    \caption{Teleoperation system setup. Solid arrows denote wired connections and dashed arrows denote wireless connections.}
    \label{fig:hardware_setup}
\end{figure}

We define end-to-end latency as the time between a motion of the human operator and the corresponding motion of the robot. As shown in Fig.~\ref{fig:hardware_setup}, each human pose passes through four stages before it is reflected in the robot motion. (1) \textit{Sensing}: the pose is captured by OptiTrack and streamed through NatNet at 120\,Hz, or captured by the VR system and published at 60\,Hz over a wireless link. (2) \textit{Transmission}: the pose is sent over a wired network to the desktop that runs the broadcaster and the policy. (3) \textit{Computation}: Direct Extremity Control maps the human pose to humanoid link targets, and the policy, running at 50\,Hz, outputs target joint positions. Mapping, policy inference, and robot communication via unitree\_sdk2\_python (1000\,Hz over CycloneDDS/UDP) run in a single Python process. At each control step, the Unitree G1 receives the PD gains $(k_p, k_d)$, the target joint position $q^*_t$, and the target joint velocity $\eta\,(q^*_t - q^*_{t-1})/\Delta t$. (4) \textit{Tracking}: the low-level PD controller drives the joints toward these targets, and the robot motion lags behind the target by the closed-loop tracking delay. Tab.~\ref{tab:latency_breakdown} lists the delay contributed by each stage.

\begin{table}[h]
\centering
\begin{tabular}{lcc}
\toprule
Component & MoCap (ms) & VR (ms) \\
\midrule
Pose streaming (average sample age) & 4 & 8 \\
Wireless transmission & -- & 8 \\
Inter-desktop communication & 0.4 & 0.4 \\
Direct extremity mapping & 0.3 & 0.3 \\
Policy inference & 0.06 & 0.06 \\
Robot communication & $<$1 & $<$1 \\
Closed-loop motion tracking & 46 & 46 \\
\midrule
Sum of components & 51 & 63 \\
Measured end-to-end & 54 & 64 \\
\bottomrule
\end{tabular}
\caption{System-level latency breakdown. Closed-loop motion tracking is measured in simulation with $\eta=0.9$ and includes the average 10\,ms wait induced by the 50\,Hz policy update.}
\label{tab:latency_breakdown}
\end{table}

The breakdown shows that end-to-end latency is dominated by closed-loop tracking. Tracking contributes 46\,ms, which accounts for about 85\% of the measured latency under MoCap and about 70\% under VR. Sensing is the second largest source: since poses arrive asynchronously, each pose has waited on average half a streaming period before it is used, which is 4\,ms at 120\,Hz and 8\,ms at 60\,Hz, and the wireless link adds another 8\,ms for VR. These two sensing terms explain the 10\,ms difference between the MoCap and VR settings. All transmission and computation steps together contribute less than 2\,ms, so the use of Python and the communication stack does not noticeably affect latency. The sum of all components is within 3\,ms of the measured latency in both settings, and we attribute the remaining difference to internal processing in the motion capture and VR systems. Since tracking delay is the only large component, reducing it is the most effective way to reduce end-to-end latency, which is exactly what scaled velocity feedforward does.

\subsection{Whole-Body Impedance Calibration}

\begin{figure}[h]
    \centering
    \includegraphics[width=0.6\textwidth]{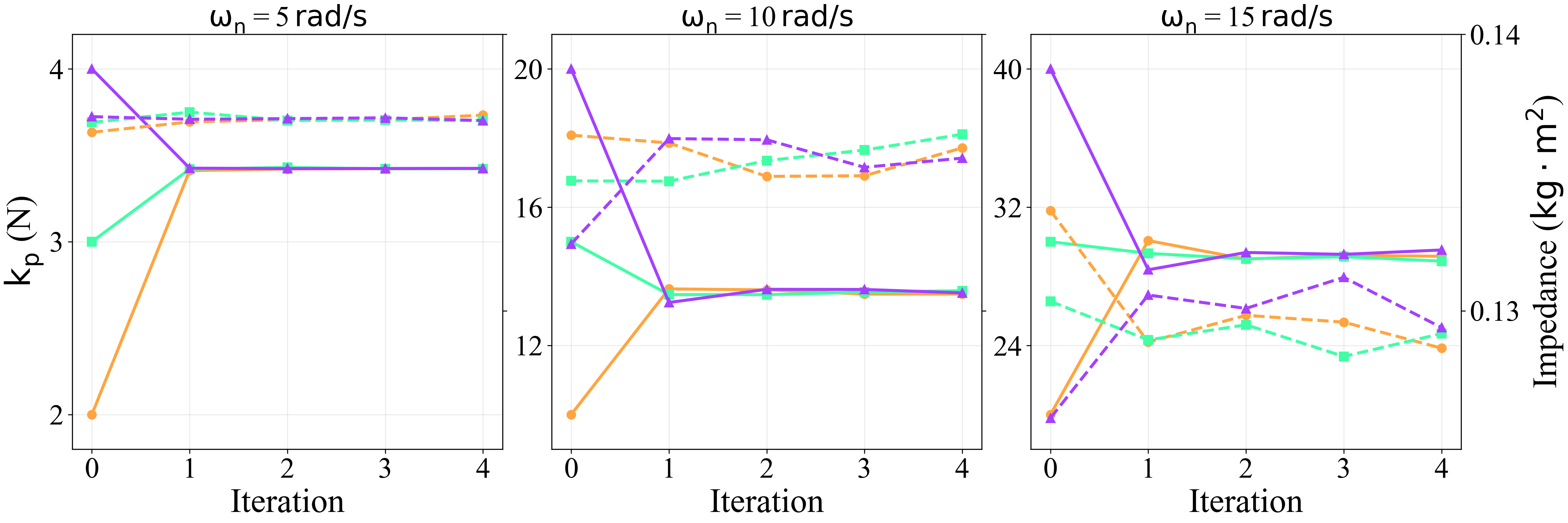}
    \caption{Whole-body impedance calibration of Unitree G1 elbow joint. Solid lines correspond to proportional gains $k_p$, dashed lines depict the effective impedance.}
    \label{fig:kp_impedance_all}
\end{figure}

 As shown in Fig.\ref{fig:kp_impedance_all}, we evaluate the proposed whole-body impedance calibration under different target natural frequencies $\omega_n$ and initial proportional gains $k_p$ on the Unitree G1 elbow joint, while three additional distal joints are calibrated simultaneously but not visualized. The derivative gain $k_d$ is computed using a fixed damping ratio $\zeta=1$ as defined in Eq.~\eqref{eq:pd_from_impedance}.
Despite large variations in initialization, the calibration process consistently converges to stable effective impedance values that differ across $\omega_n$, demonstrating robustness to initial gains and reliable recovery of $M_{\mathrm{eff}}$ induced by whole-body coupling.

\subsection{Velocity Feedforward Control}

\begin{figure}[h]
    \centering
    \includegraphics[width=0.6\textwidth]{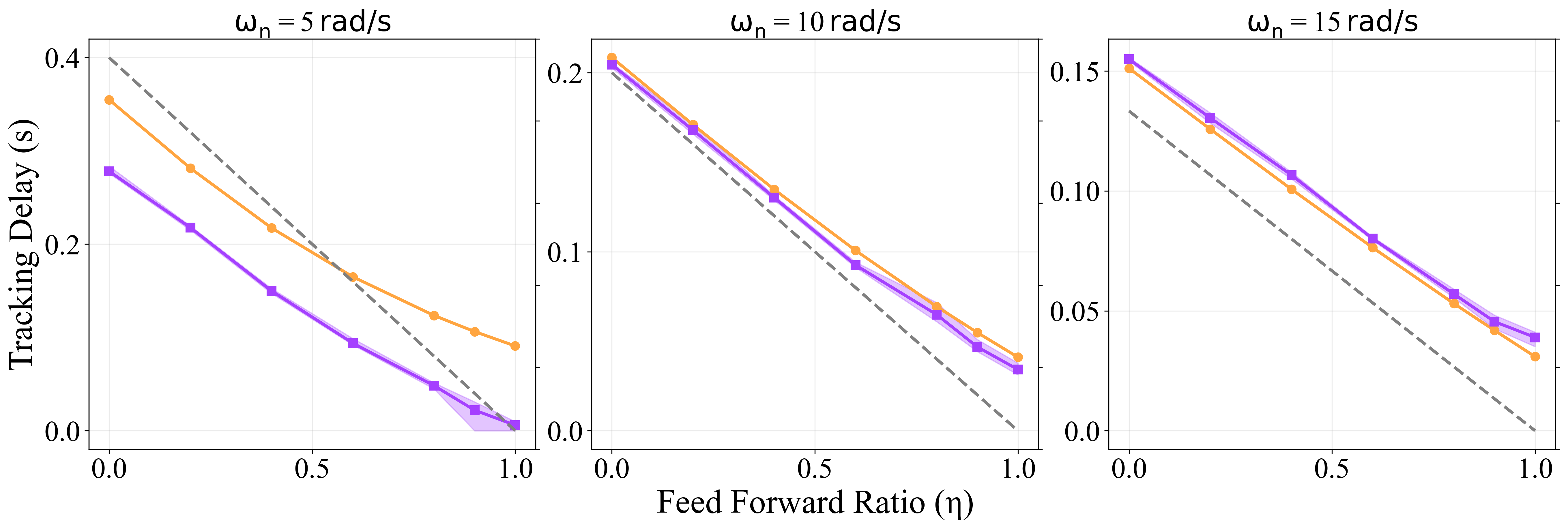}
    \caption{
    Measured tracking delay in \orange{simulation} and \purple{real world} as a function of velocity feedforward ratio. \textcolor[rgb]{0.5,0.5,0.5}{Dashed line} represents the theoretical value $\frac{2(1-\eta)}{\omega_n}$.
    }
    \label{fig:feedforward_vs_lag_sim_and_real_all}
\end{figure}

Using the PD gains obtained from whole-body impedance calibration, we evaluate the low-level tracking delay predicted by Eq.~\eqref{eq:latency} in both simulation and on hardware. A sinusoidal reference with frequency $\omega = 3.14\,\mathrm{rad}/\mathrm{s}$ is applied as the target joint position $q_t$.

As shown in Fig.~\ref{fig:feedforward_vs_lag_sim_and_real_all}, the theoretical predictions closely match the measured delays for $\omega_n = 10\,\mathrm{rad}/\mathrm{s}$ and $\omega_n = 15\,\mathrm{rad}/\mathrm{s}$, with the remaining discrepancy explained by the control period $\Delta t = 0.02\,\mathrm{s}$. For $\omega_n = 5\,\mathrm{rad}/\mathrm{s}$, deviations arise from violation of the assumption $\omega \ll \omega_n$ and numerical inaccuracies.
With a control update frequency of 50 Hz, the measured delay increases at higher velocity feedforward ratios, producing the upward trend observed at the right end of the curves. This effect is consistent with the discrete-time overshoot discussed by Eq.~\eqref{eq:upperbound}. These results validate the theoretical analysis of velocity feedforward and impedance calibration and the correctness of the low-level system design.

\subsection{Optical Flow-based Latency Estimation}
\label{sec:real_optical}

\begin{figure}[h]
    \centering
    \includegraphics[width=0.6\textwidth]{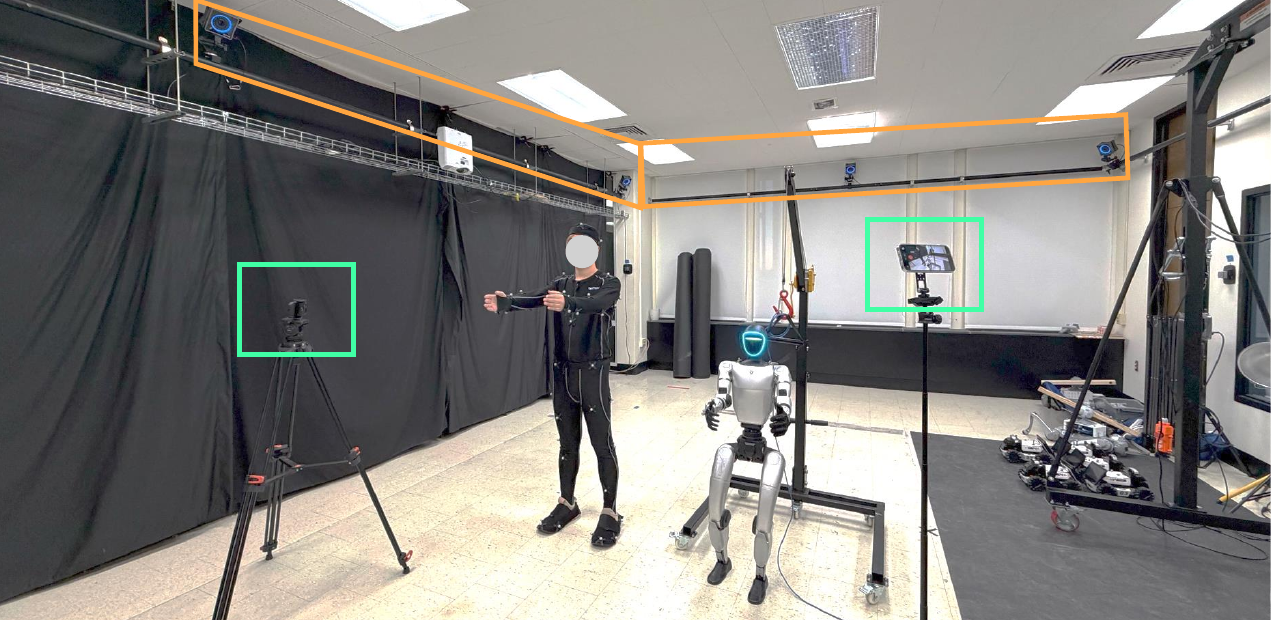}
    \caption{Experimental setup for latency validation, with \cyan{two cameras} and \orange{an optical motion capture system} simultaneously recording.}
    \label{fig:latency_setup}
\end{figure}

To demonstrate the accuracy and reliability of the proposed video-based latency estimation method, we conduct a controlled validation experiment using multi-view video recordings together with motion-capture-based ground-truth measurements. Specifically, we record a single reciprocating hand motion simultaneously using two cameras placed at different viewpoints, observing the same motion from distinct perspectives, as shown in Fig.~\ref{fig:latency_setup}. For each camera view, we independently apply the optical-flow-based pipeline described in Sec.~\ref{sec:experiment} to extract a one-dimensional motion signal and estimate the temporal offset between the human and robot motions, yielding two latency estimates.

\begin{figure}[h]
    \centering
    \includegraphics[width=0.8\textwidth]{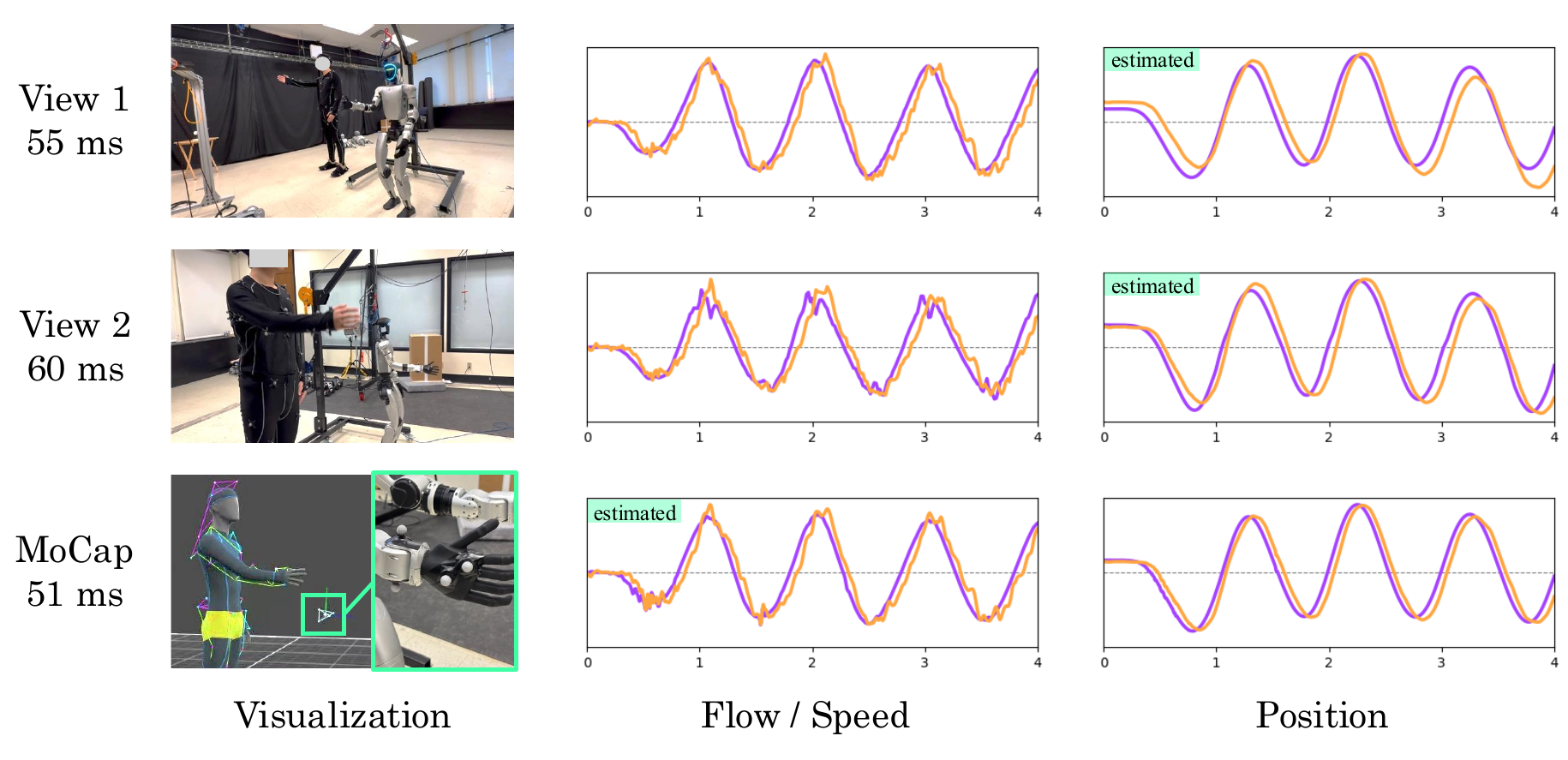}
    \caption{Normalized optical-flow projections and the corresponding position trajectories of the \purple{human} and \orange{robot} are shown on the right.}
    \label{fig:latency_supp}
\end{figure}

Fig.~\ref{fig:latency_supp} presents the experimental results. For the two camera views, the middle column visualizes the projected optical flows, while the right column shows the accumulated displacement obtained by integrating the projected flow over time, which approximates the underlying motion trajectory. \textbf{The estimated latencies are 55\,ms and 60\,ms}, and their agreement serves as a consistency check for view invariance and robustness of the method. For the motion-captured end-effector positions, the middle column compares the estimated velocities, and the right column shows the directly tracked position waveforms. \textbf{The ground-truth latency calculated directly from the tracked positions is 51\,ms}, which, together with the velocities computed from these positions, is consistent with the optical-flow-based estimates, confirming that the proposed method accurately captures end-to-end system latency. Minor discrepancies may arise due to non-ideal camera viewpoints or imperfect alignment between the chosen projection direction and the true motion. As illustrated in Fig.~\ref{fig:latency_supp}, the projected optical flow in View~1 closely aligns with the motion-capture-derived velocity, whereas the flow in View~2 is noticeably noisier, leading to a less accurate latency estimate. In addition, sub-frame-level variance (less than $16$\,ms for $60$\,FPS videos) can naturally occur due to measurement noise and temporal discretization effects.

\newpage
\section{Policy Learning Details}

\label{sec:policy_learning}

To clarify the policy learning formulation, we first introduce the notation used in this section. As defined in Sec.~\ref{sec:method_retarget}, we denote the target pose of each tracking link as $\mathbf{T}^r_{\mathrm{link}} = [\mathbf{R}^r_{\mathrm{link}}, \mathbf{p}^r_{\mathrm{link}}] \in \mathrm{SE}(3)$, and use $\mathbf{T}^r$ to represent the collection of all tracked link poses. The pose discrepancy between the measured and target link poses is denoted as $\Delta \mathbf{T}^r_{\mathrm{link}} = [\Delta \mathbf{R}^r_{\mathrm{link}}, \Delta \mathbf{p}^r_{\mathrm{link}}]$.
Notably, the pelvis link serves as the root link in simulation and as the IMU mounting base in the real robot; therefore, it is used to represent the global position and orientation of the robot. 

Based on the foot link position $\mathbf{p}^r_{k,\mathrm{foot}}$, we estimate the probability of foot–ground contact as follows:
\[
\mathbb{P}_\mathrm{k\_foot}=1-\min(1, \frac{\mathbf{p}^r_\mathrm{k\_foot,z}-0.2}{0.2}+\frac{||\dot{\mathbf{p}}^r_\mathrm{k_foot,xy}||-0.2}{0.2})
\]
Based on the estimated contact probability and the principle of momentum conservation, we approximate the contact force. We denote $\mathbb{F}$ as the proportion of the total body weight supported by each contact.
\[
\mathbb{F}^t_\mathrm{k\_foot}=\mathbb{P}^t_\mathrm{k\_foot}\big/\mathbb{E}[\mathbb{P}^{t-15:t+15}_\mathrm{left\_foot}+\mathbb{P}^{t-15:t+15}_\mathrm{right\_foot}]
\]
where $\mathbb{E}$ denotes a weighted sum with a quadratic weighting function that vanishes at both endpoints.

\subsection{Problem Formulation}
\label{sec:problem_formulation}

We model the control problem as a Markov Decision Process $(\mathcal{O}, \mathcal{A}, \mathcal{P}, \mathcal{R}, \gamma)$, where the policy $\pi^*: \mathcal{O} \rightarrow \mathcal{A}$ maps observations $\mathbf{o}_t \in \mathbb{O}$ to actions $\mathbf{a}_t \in \mathbb{A}$, interpreted as scaled joint target positions $\mathbf{q}^*_t = k \cdot \mathbf{a}_t$ for a fixed scalar $k \in \mathbb{R}_{>0}$. Teleoperation is formulated as a one-step-ahead motion tracking problem, where the operator provides end-effector target poses $\mathbf{T}_t \in SE(3)$ one timestep in advance, and the policy minimizes the tracking error $||\mathbf{T_t}-\mathbf{T}_{t-1}^r||$
where $\|\cdot\|$ denotes an appropriate $SE(3)$ distance metric.

\subsection{Observation}

Since our control interface operates solely on a selected set of robot link poses, we introduce a future interpretation function to encode reference motion over a finite horizon $H$
\begin{equation}
\mathbf{o}^{\mathrm{ref}}(H)=\mathscr{I}(\mathbf{T}^r_{t:t+H}, \mathbf{V}^r_{t:t+H},\mathbf{T}^r_{t-1})
\end{equation}
where $\mathbf{V}^r=\dot{\mathbf{T}}^r(\mathbf{T}^r)^{-1}$ represents the target velocities in the global frame. $\mathscr{I}$ is defined as follow: 

\begin{table}[h]
\centering
\vspace{-5pt}
\begin{threeparttable}
\setlength{\tabcolsep}{6pt}
\renewcommand{\arraystretch}{1.4}
\setlength{\aboverulesep}{2pt}
\setlength{\belowrulesep}{-1pt}

\begin{tabular*}{0.76\textwidth}{l|l}
\toprule
\\[-11.5pt]
Future pelvis translation &
$\mathbf{p}^r_\mathrm{pelvis,t:t+H}-\mathbf{p}^r_\mathrm{pelvis,t-1}$
\\
\hline
Future pelvis rotation &
$\mathbf{r}^{r,\top}_\mathrm{pelvis,t:t+H}\cdot\mathbf{r}^r_\mathrm{pelvis,t-1}$
\\
\hline
Link poses &
$\mathbf{T}^{r'}_{t:t+H}$
\\
\hline
Link velocities (privilege) &
$\mathbf{V}^r_{t:t+H}$
\\
\hline
Retargeted joint configuration (privilege) &
$q_t,\dot{q}_t$
\\
\hline
Foot contact probability &
$\mathbb{P}_\mathrm{k\_foot}$
\\[-2pt]
\bottomrule

\end{tabular*}
\end{threeparttable}
\vspace{-10pt}
\end{table}

To explicitly capture extremity pose errors and account for the IMU being mounted on the pelvis of the Unitree G1, we express both target poses ${\mathbf{T}^r}'$ and actual poses ${\mathbf{U}^r}'$ in the pelvis frame, where ${\mathbf{U}^r}'$ is obtained via forward kinematics from the pelvis, and compute the local pose discrepancy as
\begin{equation}
\mathbf{o}^{\mathrm{diff}}=({\mathbf{U}^r}')^{-1}{\mathbf{T}^r}'.
\end{equation}
 For deployable policies $\pi_{\mathrm{student}}$ and $\pi_{\mathrm{teleop}}$, we use a single future frame to maximize responsiveness, resulting in the observation vector
$[\mathbf{o}^\mathrm{ref}(1),\mathbf{o}^\mathrm{diff},\mathbf{o}^\mathrm{proprio}]$ where $\mathbf{o}^\mathrm{proprio}$ denotes proprioceptive observations. The oracle policy $\pi_{\mathrm{teacher}}$ leverages a longer horizon together with privileged observations $\mathbf{o}^\mathrm{priv}$, including global-frame pose discrepancies, joint configurations after inverse kinematics, domain randomization parameters and foot contact forces, yielding $[\mathbf{o}^\mathrm{ref}(32),\mathbf{o}^\mathrm{diff},\mathbf{o}^\mathrm{proprio},\mathbf{o}^\mathrm{priv}]$.
\newpage

The proprioception observation $\mathbf{o}^\mathrm{proprio}$ is listed below:

\begin{table}[h]
\centering
\vspace{-5pt}
\begin{threeparttable}
\setlength{\tabcolsep}{6pt}
\renewcommand{\arraystretch}{1.4}
\setlength{\aboverulesep}{2pt}
\setlength{\belowrulesep}{-1pt}

\begin{tabular*}{0.46\textwidth}{l|l}
\toprule
\\[-11.5pt]
Last action &
$a_{t-1}$
\\
\hline
Joint configuration &
$q,\dot{q}$
\\
\hline
Pelvis rotation &
$\mathbf{r}_\mathrm{pelvis}$
\\
\hline
Pelvis linear velocity (privilege) &
$\dot{\mathbf{p}}_\mathrm{pelvis}$
\\
\hline
Pelvis angular velocity &
$\dot{\mathbf{r}}_\mathrm{pelvis}$
\\[-2pt]
\bottomrule

\end{tabular*}
\end{threeparttable}
\vspace{-5pt}
\end{table}

Additional privileged observation $\mathbf{o}^\mathrm{priv}$ is listed below:

\begin{table}[h]
\centering
\vspace{-5pt}
\begin{threeparttable}
\setlength{\tabcolsep}{6pt}
\renewcommand{\arraystretch}{1.4}
\setlength{\aboverulesep}{2pt}
\setlength{\belowrulesep}{-1pt}

\begin{tabular*}{0.6\textwidth}{l|l}
\toprule
\\[-11.5pt]
Domain Randomization Parameters &
---
\\
\hline
Residual joint configuration &
$\Delta q,\Delta\dot{q}$
\\
\hline
Residual link poses &
$\Delta\mathbf{T}^r$
\\
\hline
Foot contact force &
$\mathbb{F}^\mathrm{target}_\mathrm{k\_foot}, \mathbb{F}^\mathrm{measure}_\mathrm{k\_foot}$
\\[-2pt]
\bottomrule

\end{tabular*}
\end{threeparttable}
\vspace{-5pt}
\end{table}

\subsection{Reward Function}

To improve exploration stability in online reinforcement learning, we formulate all reward terms as negative-valued and compute the exponential of their weighted sum as the total reward. This formulation ensures that (i) the per-step total reward lies within $(0, 1)$, and (ii) the gradient of the total reward with respect to each individual term is preserved. The complete specification of the reward functions is provided in the appendix.

The exact reward formulation involves axis- and link-specific weighting terms and is therefore omitted for brevity. Below, we present the simplified primary reward components.

\begin{table}[h]
\centering
\vspace{-5pt}
\begin{threeparttable}
\setlength{\tabcolsep}{6pt}
\renewcommand{\arraystretch}{1.4}
\setlength{\aboverulesep}{2pt}
\setlength{\belowrulesep}{-1pt}

\begin{tabular*}{0.745\textwidth}{l|l|l}
\toprule
\\[-13.5pt]
\rowcolor{gray!15}
\textbf{Reward Term} &
\textbf{Expression} & \textbf{Scale}
\\
\hline
Global tracking link poses &
$-||(\mathbf{U}^r)^{-1}\mathbf{T}^r||_2^2$ & 30
\\
\hline
Local tracking link poses &
$-||(\mathbf{U}^{r'})^{-1}\mathbf{T}^{r'}||_2^2$ & 20
\\
\hline
Retargeted DoF position &
$-||q_t-q||_2^2$ & 3
\\
\hline
Retargeted DoF velocity &
$-||\dot{q}_t-\dot{q}||_2^2$ & 0.02
\\
\hline
Foot contact reward &
$-\left(\mathbb{F}^\mathrm{target}_\mathrm{k\_foot}-\mathbb{F}^\mathrm{measure}_\mathrm{k\_foot}\right)^2$ & 3
\\
\hline
Foot contact penalty &
$-\left(\mathbb{I}[\mathbb{P_\mathrm{k\_foot}}<0.2]\cdot\mathbb{F}^\mathrm{measure}_\mathrm{k\_foot}\right)^2$ & 10
\\
\hline
Torque Penalty &
$-||\tau||_2^2$ & 0.0001
\\
\hline
Action rate &
$-||a_{t-1}-a_{t}||_2^2$ & 0.2
\\[-2pt]
\bottomrule

\end{tabular*}
\end{threeparttable}
\vspace{-5pt}
\end{table}

\subsection{Motion Dataset}

As observed in prior work \cite{TWIST,TWIST2}, incorporating in-domain motion data significantly improves policy performance during teleoperation. We collect a set of motion sequences $\mathcal{S}_\mathrm{teleop}$ using an optical motion capture system; however, to ensure fair evaluation and reproducibility, these user-specific datasets are excluded from policy training experiments unless explicitly stated. 

Aiming for task-agnostic policy learning, we use the widely adopted LAFAN1 dataset $\mathcal{S}_\mathrm{lafan}$~\cite{harvey2020robust} retargeted by Unitree~\cite{lv2025lafan1_retargeting}, as dynamic and challenging motions, and AMASS $\mathcal{S}_\mathrm{amass}$~\cite{mahmood2019amass} as a large-scale and diverse motion corpus. Unless otherwise specified, we use $\mathcal{S}_\mathrm{lafan}$ in teacher and student learning stages, $\mathcal{S}_\mathrm{lafan}\cup\mathcal{S}_\mathrm{amass}$ in RL finetune stage. For evaluation, we additionally collect a trajectory $\mathcal{S}_\mathrm{teleop}'=\{s_\mathrm{eval}\},s_\mathrm{eval}\notin\mathcal{S}_\mathrm{teleop}$ comprising common daily motions to assess teleoperation performance.

\newpage
\subsection{Domain Randomization} Domain randomization is applied to motor dynamics, contact friction, and torso mass distribution. For motor dynamics, we apply four randomizations:
\[
\tau=\alpha_\mathrm{strength}(\alpha_{k_p}k_p(q_t-q+\beta_\mathrm{offset})+\alpha_{k_d}(\eta\,\dot{q}_t-q_t))
\]

Detailed distributions are as follows:

\begin{table}[h]
\centering
\vspace{-5pt}
\begin{threeparttable}
\setlength{\tabcolsep}{6pt}
\renewcommand{\arraystretch}{1.4}
\setlength{\aboverulesep}{2pt}
\setlength{\belowrulesep}{-1pt}

\begin{tabular*}{0.53\textwidth}{l|l}
\toprule
\\[-11.5pt]
$k_p$ ratio $\alpha_{k_p}$ &
$\mathcal{U}(0.8,1.2)$
\\
\hline
$k_d$ ratio $\alpha_{k_d}$&
$\mathcal{U}(0.8,1.2)$
\\
\hline
Motor strength $\alpha_\mathrm{strength}$ &
$\mathcal{U}(0.8,1.2)$
\\
\hline
Motor offset $\beta_\mathrm{offset}$ &
$\mathcal{U}(-0.1,0.1)$
\\
\hline
Friction ratio &
$\mathcal{U}(0.3, 1.0)$
\\
\hline
Torso added mass (kg) &
$\mathcal{U}(-2, 5)$
\\
\hline
Torso CoM displacement (m) &
$\mathcal{U}(-0.05, 0.05)^3$
\\[-2pt]
\bottomrule

\end{tabular*}
\end{threeparttable}
\vspace{-5pt}
\end{table}

\subsection{Hyperparameters} All actor and critic networks are implemented as MLPs with hidden dimensions [1024, 512, 256] and ReLU as activation layer. We adopt an adaptive learning rate schedule in PPO with a target KL divergence of 0.01. The remaining PPO hyperparameters are listed below:

\begin{table}[h]
\centering
\vspace{-5pt}
\begin{threeparttable}
\setlength{\tabcolsep}{6pt}
\renewcommand{\arraystretch}{1.4}
\setlength{\aboverulesep}{2pt}
\setlength{\belowrulesep}{-1pt}

\begin{tabular*}{0.35\textwidth}{l|l}
\toprule
\\[-11.5pt]
$\gamma$ & 0.99
\\
\hline
$\lambda_\mathrm{GAE}$ & 0.95
\\
\hline
Entropy coefficient & 0.003
\\
\hline
Value loss coefficient & 1.0
\\
\hline
Rollout length & 24
\\
\hline
Optimizer & ADAM
\\
\hline
$\#$ epoch & 5
\\
\hline
$\#$ mini batch & 8
\\
\hline
$\#$ iteration & 6000
\\[-2pt]
\bottomrule

\end{tabular*}
\end{threeparttable}
\vspace{-5pt}
\end{table}

We utilize DAgger with the following hyperparameters:

\begin{table}[h]
\centering
\vspace{-5pt}
\begin{threeparttable}
\setlength{\tabcolsep}{6pt}
\renewcommand{\arraystretch}{1.4}
\setlength{\aboverulesep}{2pt}
\setlength{\belowrulesep}{-1pt}

\begin{tabular*}{0.29\textwidth}{l|l}
\toprule
\\[-11.5pt]
learning rate $\eta$ & 0.0003
\\
\hline
Batch size & 256
\\
\hline
Rollout length & 24
\\
\hline
Optimizer & ADAM
\\
\hline
$\#$ epoch & 10
\\
\hline
$\#$ iteration & 1500
\\[-2pt]
\bottomrule

\end{tabular*}
\end{threeparttable}
\vspace{-5pt}
\end{table}

\end{document}